\documentclass[]{iopjournal}

\ifdefined\pdfvariable
  \pdfvariable minorversion 7
\else
\fi

\usepackage[T1]{fontenc}
\usepackage{lmodern}
\usepackage{bm}
\usepackage{cite}
\usepackage{flushend}
\usepackage{siunitx}
\usepackage{multirow}
\usepackage{amsmath}
\usepackage{amsfonts}
\usepackage{xcolor}
\usepackage{placeins}

\DeclareMathOperator*{\argmax}{arg\,max}

\newcommand{\tabref}[1]{{Table \ref{#1}}}
\newcommand{\figref}[1]{{Fig. \ref{#1}}}
\newcommand{\equref}[1]{{Eq. \ref{#1}}}

\fancyhfoffset[L]{0pt}
\hypersetup{
  pdftitle={Robotic Physical Comparison of Biologically Motivated and Counterfactual Sensory Arrangements using Biomimetic Skin},
  pdfauthor={Akihiro Miki, Shun Hasegawa, Yoshimoto Ribayashi, Yuta Sahara, Kento Kawaharazuka, Kei Okada}
}

\begin{document}

\title
  {
      Robotic Physical Comparison of Biologically Motivated and Counterfactual Sensory Arrangements using Biomimetic Skin
  }

\author{Akihiro Miki$^1$\orcid{0009-0006-1934-1405},
Shun Hasegawa$^1$\orcid{0000-0002-7443-2987},
Yoshimoto Ribayashi$^1$\orcid{0009-0000-8606-0896},
Yuta Sahara$^1$\orcid{0009-0007-3103-7243},
Kento Kawaharazuka$^{1,2}$\orcid{0000-0002-7464-7187} and
Kei Okada$^1$\orcid{0000-0001-6606-6692}}

\affil{$^1$ Department of Mechano-Informatics, Graduate School of Information Science and Technology, The University of Tokyo, 7-3-1 Hongo, Bunkyo-ku, Tokyo, 113-8656, Japan.}

\affil{$^2$ AI Center, Graduate School of Information Science and Technology, The University of Tokyo, Japan.}

\email{[miki, hasegawa, ribayashi, sahara, kawaharazuka, k-okada]@jsk.t.u-tokyo.ac.jp}


\par\medskip
\begingroup\footnotesize
\noindent\textbf{Publication information.}
A. Miki, S. Hasegawa, Y. Ribayashi, Y. Sahara, K. Kawaharazuka and K. Okada,
\textit{Bioinspiration \& Biomimetics} \textbf{21}(4), 046009 (2026).
This is the Accepted Manuscript version of an article accepted for publication in
Bioinspiration \& Biomimetics. IOP Publishing Ltd is not responsible for any errors
or omissions in this version of the manuscript or any version derived from it.
This Accepted Manuscript is published under a
\href{https://creativecommons.org/licenses/by/4.0/}{CC BY 4.0 licence}.
The Version of Record is available online at
\href{https://doi.org/10.1088/1748-3190/ae7786}{https://doi.org/10.1088/1748-3190/ae7786}.
\par\endgroup

\begin{abstract}

      Human skin contains diverse mechanoreceptors, including Merkel cells and Pacinian corpuscles, each localized in specific regions.
      Although their spatial organization is known to be shaped by molecular signaling and neural induction during development, experimentally examining its functional significance under controlled conditions in living systems remains challenging.
      In particular, physically comparing alternative receptor arrangements that do not occur biologically is difficult.
      In this study, we developed a layered biomimetic robotic skin designed to mimic key structural features of human skin tissue by embedding receptor-analog sensors within compliant soft media.
      Using this platform, we implemented both biologically motivated and counterfactual sensory arrangements and quantitatively evaluated their performance across three passive tactile tasks: detection of weak contacts, a two-point discrimination-like assessment, and texture identification.
      Arrangements reflecting key features of the biologically motivated sensory organization exhibited higher performance than the alternative configurations examined.
      This study addresses a biological question concerning how spatial organization influences tactile function by presenting a biomimetic robotic approach that enables physical and empirical comparison of biologically motivated and counterfactual sensory arrangements that are difficult to investigate in living systems.

\end{abstract}

\section{Introduction}\label{sec:self_organized_skin_introduction}
  The skin is composed of three layers: epidermis, dermis, and subcutaneous tissue (hypodermis) \cite{Lotfollahi:2024:SkinAnatomy}.
  The epidermis, the outermost layer, consists mainly of keratinocytes and serves as a protective barrier against the external environment.
  Beneath it lies the dermis, which contains connective tissues such as collagen and elastin.
  The deeper subcutaneous tissue comprises adipocytes, blood vessels, and nerves, contributing to thermoregulation and absorption of mechanical impact.

  Within these skin tissues, a variety of sensory receptors—including Merkel cells, Pacinian corpuscles, Ruffini endings, and Meissner corpuscles—are distributed at different depths and regions depending on the nature of the stimulus.
  These receptors enable precise detection of pressure, vibration, slip, and skin stretching.
  For example, Merkel cells are located in the basal layer of the epidermis and respond to sustained pressure \cite{Ogawa:1996:MerkelCell}.
  Meissner corpuscles reside at the dermal–epidermal junction and are sensitive to light touch and slip \cite{Cauna:1956:MeissnerCorpusclesNerveSupply}.
  Ruffini endings, found in the dermis, are thought to detect skin stretching, although morphological evidence for their structure remains limited \cite{Chambers:1972:RuffiniEndingsStructure, Olson:2016:lowthresholdMechanoreceptors, Cobo:2021:HumanDigitalRuffini}.
  Pacinian corpuscles are distributed from the deep dermis into the subcutaneous tissue and respond to high-frequency vibrations with a peak sensitivity approximately 200 Hz \cite{Loewenstein:1965:PacheconCorpuscleAdaptation}.
  Collectively, these receptors are arranged according to their functional properties at distinct depths and regions within the skin, thereby enabling the integration of diverse tactile information \cite{Johnson:2001:FourTypeMechanoreceptorsRoles, Abraira:2013:SensoryNeuronsTouch}.

  Studies in developmental biology and neuroscience have demonstrated the involvement of molecular signals and neural induction in the formation of sensory receptor arrangements.
  For instance, Merkel cells are epithelial cells that differentiate from a subset of the epidermis during embryogenesis, and their patterning is regulated by the spatial and temporal actions of diverse developmental signals such as Wnt, Shh, and BMP \cite{Xiao:2016:MerkelCellDevelopment, Jenkins:2019:MerkelCellInnervationDevelopment}.
  The formation of Pacinian corpuscles also involves multiple molecular factors, among which the GDNF family ligand neurturin (NRTN) and its receptors $\text{GFR}\alpha2/\text{Ret}$ play a critical role \cite{Luo:2009:RetSignaling}.
  In addition, several neurotrophins—including nerve growth factor (NGF), brain-derived neurotrophic factor (BDNF), and neurotrophin-3 (NT3)—have overlapping roles, as evidenced by reduced numbers of Pacinian corpuscles in knockout mice for each of these factors \cite{Vsedy:2004:PacinianCorpuscleDevelopment}.
  Collectively, these findings indicate that diverse molecular signals, acting through interactions between skin cells and the nervous system, contribute to the formation of sensory receptors, and the mechanisms underlying the construction of tactile function are gradually being elucidated \cite{Jenkins:2017:DevelopingSenseTouch}.
  Moreover, recent studies have shown that the Piezo2 channel mediates coordinated mechanotransduction between Merkel cells and sensory neurons, further advancing our understanding of the mechanisms underlying mechanosensory neuron activation \cite{Woo:2014:Piezo2ForMerkelCellMechanotransduction, Handler:2021:MechanosensoryNeuronsActivation}.

    Despite substantial progress in understanding the developmental mechanisms underlying mechanoreceptor formation, it remains unclear how the spatial organization of these receptors influences tactile function.
    In particular, because receptor configurations are largely fixed during development, it is experimentally difficult to systematically compare alternative arrangements in living systems.
    To address this limitation, this study proposes a biomimetic robotic framework that enables physical comparison between biologically motivated and counterfactual sensory arrangements by mimicking the layered structure of skin at the tissue level.
    Here, “biologically motivated” refers to configurations designed to preserve qualitative organizational principles observed in biological receptor arrangements under physical constraints, rather than strict quantitative anatomical reproduction.
    Specifically, we construct a biomimetic skin in which receptor-like sensors can be embedded in different configurations, thereby providing an experimental platform that allows comparative investigation of functional consequences, including configurations that are difficult to isolate and evaluate in biological systems.
    This study aims to present an approach that enables empirical investigation, from a biomimetic robotics perspective, of how differences in spatial organization influence tactile function under controlled conditions.

    Within this experimental framework, this study adopts physical experiments rather than simulation-based approaches.
    This is because tactile phenomena such as sliding interactions involving stick–slip dynamics are difficult to model accurately.
    In addition, another reason is that this study emphasizes a constructive approach in robotics, which assumes that certain system-level properties may only become apparent through physical implementation and experimental validation.

    The skin is an organ that covers the entire body, from the trunk to the hands and plantar surfaces.
    Although the density and functional properties of mechanoreceptors vary across different regions, several major types of mechanoreceptors are distributed across multiple skin areas and exhibit characteristic spatial organization corresponding to the layered structure of the skin.
    In this study, we focus on representative mechanoreceptors that are present in both trunk and distal regions, namely Merkel cell complexes and Pacinian corpuscles, and examine their spatial arrangements to discuss aspects of sensory organization in the skin.

  Numerous approaches have been developed in robotics and haptics research to achieve highly accurate tactile sensing through the integration of various technologies.
  For example, BioTac, a fingertip sensor inspired by human tactile function \cite{Fishel:2012:BioTac}, consists of a flexible elastomeric skin filled with conductive fluid and equipped with multiple electrodes on its surface, and has been applied to studies of object recognition \cite{Fishel:2012:BioTacBayesian, Chu:2015:BioTacHapticAdjectives}.
  Other examples include fingertip sensors embedded with strain gauges and PVDF films in random distributions \cite{Hosoda:2006:AnthropomorphicFinger}, and designs that place force sensors in deeper layers and accelerometers in more superficial layers \cite{Chathuranga:2012:BiomemticFingertip}.
  Advances have also been made in modular skin systems designed to cover the entire body of a robot, including conformable modules \cite{Ohmura:2006:ConformableSkin} and large-scale tactile sensing arrays composed of interconnected triangular or hexagonal modules \cite{Maiolino:2013:LargeScaleCapacitiveTactileSystem, Cheng:2019:ComprehensiveSkin}.
  More recently, research on flexible sensors based on novel materials such as conductive polymers, gels, and liquid metals has accelerated \cite{Wei:2019:CarbonConductivePasteSensor, Shen:2021:CutaneousIongelESkin, Won:2021:LiquidMetalWearableElectronicsReview}, and new sensing modalities, including electrical impedance and acoustic tomography, have also been proposed \cite{Park:2022:BiomimeticElastomericRobotSkin}.

  However, most of these studies have focused on achieving advanced tactile functions or improving technical performance, and few have strictly mimicked the spatial arrangement of sensory receptors found in biological skin.
  As a result, it has been difficult to deepen our understanding of biological sensory systems in ways that depend on how information changes with receptor placement.
  For example, impedance tomography techniques, which measure electrical data between multiple electrodes, reconstruct the shape deformations of the skin but do not directly capture forces or strains at specific depths.
  In contrast, this study emphasizes the spatial arrangement of the receptors and their corresponding mechanoreceptive properties.
  We constructed a biomimetic skin that incorporates not only biological structures but also modified arrangements based on them, with the goal of advancing our understanding of sensory systems.
  In particular, the fabricated biomimetic skin was quantitatively evaluated through tasks such as discrimination analogous to the two-point threshold, detection of fine pressure contacts, and recognition of different textures, thereby empirically verifying the impact of sensory receptor arrangements on tactile function.
  Previous studies have often focused on the characteristics or developmental mechanisms of single types of receptors, and very few have addressed the functional significance of arrangements comprising multiple receptor types.
  In contrast, this study considered the spatial distribution of multiple receptor types and comprehensively examines their functional effects.

    This study is expected to provide insights from a biomimetic robotics perspective for the design of multimodal sensing systems by emulating and examining biologically motivated sensory arrangements that are thought to support interaction with the environment.
    In addition, this work presents a biomimetic robotic approach to biological questions, in which physical models are used instead of living systems to enable comparative investigation not only of biologically motivated configurations but also of counterfactual configurations that are difficult to realize in vivo.
    Such an approach may offer a constructive perspective for exploring functional constraints and possible adaptive factors underlying body organization, thereby contributing to a deeper understanding of sensory arrangement in human skin and facilitating interdisciplinary research.

\section{Material and Methods} \label{sec:self_organized_skin_material_methods}
  In this study, we constructed a flexible skin structure that mimics the spatial arrangements of biological sensory receptors and conducted a series of experiments to quantitatively evaluate their effects on tactile function.
  This section first describes the design of the skin structure and sensor arrangements, followed by the experimental settings for different tactile discrimination tasks, including the robot used, data acquisition and processing, and evaluation procedures.
  The discrimination tasks targeted in this study focused on sensory receptors that are common across the skin, rather than those specialized for specific regions, such as the fingertips.
  Accordingly, we investigated passive tactile stimulation, such as pressing and stroking, rather than active exploratory behavior.
  For pressure stimulation, it is known that humans have difficulty reporting pressure intensity numerically, yet they can reliably perceive the location of contact (e.g., whether it is at one or two points) and even detect extremely light touches.
  Therefore, inspired by the classical two-point discrimination test, we examined the ability of the biomimetic skin to discriminate between one-point and two-point stimuli applied at different separations, as well as its sensitivity to very light contacts.
  In addition, since humans are known to be able to identify material properties to some extent through stroking stimuli, we also evaluated the performance of the biomimetic skin in discriminating different textures.
  Taken together, the aim of this study was to clarify how differences in sensory receptor arrangements affect tactile discrimination performance in the constructed biomimetic skin, through three types of passive tactile tasks: (i) a two-point discrimination-like test, (ii) fine-contact detection under low-pressure stimuli, and (iii) texture discrimination.

\subsection{Design and Implementation of the Biomimetic Skin} \label{subsec:self_organized_skin_construction_methods}
  In this section, we describe a method for constructing a biomimetic skin using robotic techniques and the design principles of sensory receptor arrangements.
  In biological skin, sensory receptors are primarily mechanoreceptors that respond to stimuli transmitted through the skin tissues.
  To understand the organization of these receptors, it is essential to reproduce the layered structures of tissues through which mechanical stimuli propagate.
  Accordingly, in this study, we constructed biomimetic skin at the tissue level using rubber and silicone, with an emphasis on replicating the layered architecture and elastic variations of biological skin.
  Within this artificial skin, sensory elements were embedded not only in biomimetic configurations resembling biological arrangements, but also in modified arrangements, thereby enabling deeper investigation of sensory system organization.

\begin{figure}[htbp]
  \centering
  \includegraphics[width=1.0\linewidth]{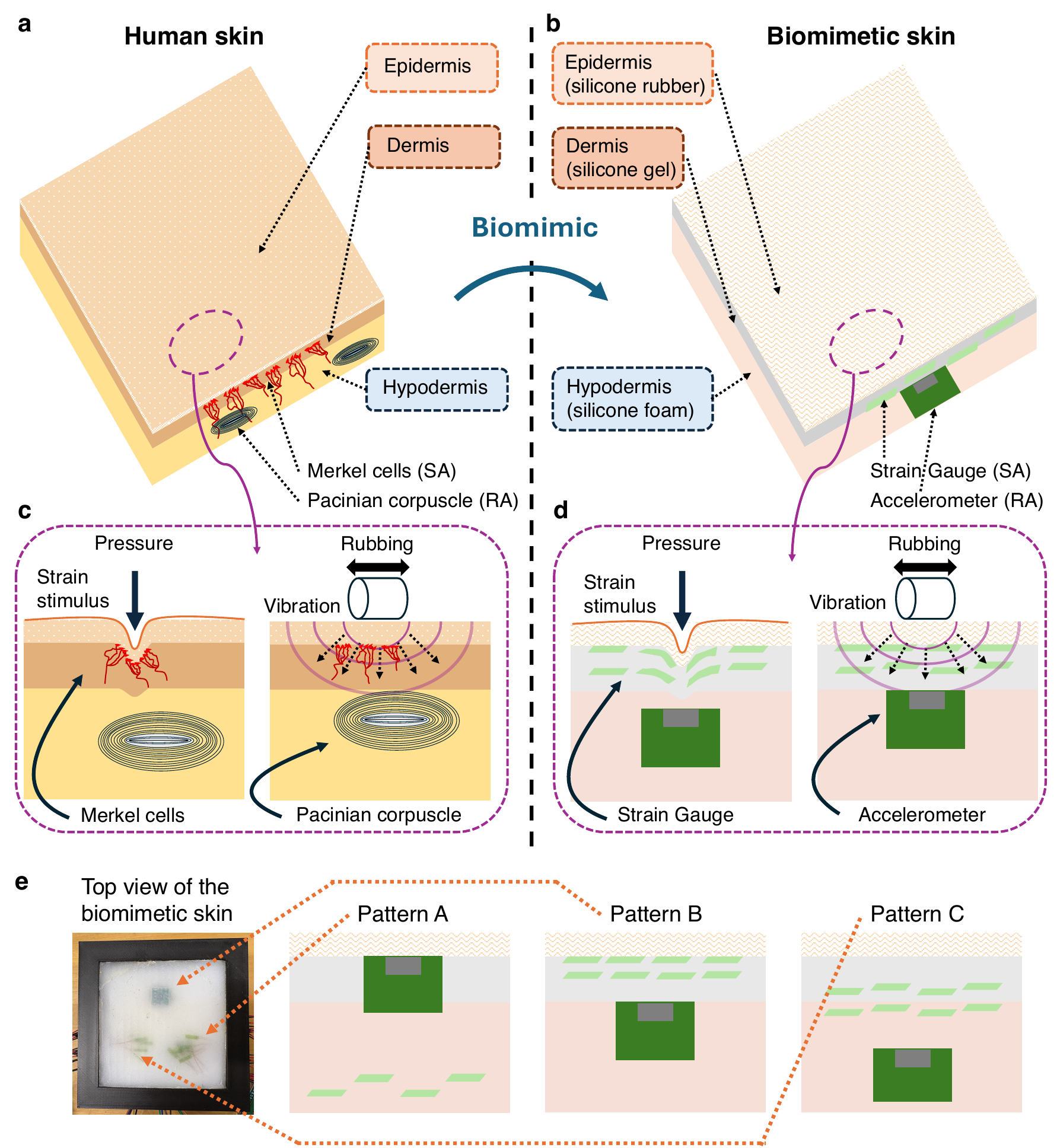}
  \caption{
    Structural and functional overview of the biomimetic skin system designed to replicate human tactile sensing.
      SA and RA denote slowly adapting and rapidly adapting responses, respectively.
      In human skin, Merkel cells exhibit SA responses, whereas Pacinian corpuscles exhibit RA responses.
      In this study, strain gauge sensors correspond to SA units, while accelerometers correspond to RA units.
    (a) \textbf{Structure of human skin:}
    A schematic diagram showing the layered anatomy of human skin—epidermis, dermis, and hypodermis—as well as mechanoreceptors (e.g., Merkel cells and Pacinian corpuscles) embedded at different depths.
    (b) \textbf{Structure of the biomimetic skin:}
    The biomimetic skin is constructed from soft, elastic materials, layered to mimic the structure of the epidermis, dermis, and hypodermis.
    Strain gauge sensors and accelerometers are embedded to replicate the distribution and depth of biological tactile receptors.
    As shown in (e), three different sensor embedding patterns were implemented for comparative analysis.
    (c) \textbf{Tactile sensing in human skin:}
    Static pressure is mainly detected by densely distributed Merkel cells, while rubbing stimuli are primarily sensed by Pacinian corpuscles, which respond to high-frequency vibrations.
    (d) \textbf{Tactile sensing in the biomimetic skin:}
    Distributed strain gauges emulate the function of Merkel cells to detect static pressure, while embedded accelerometers, mimicking Pacinian corpuscles, respond to vibratory stimuli caused by rubbing.
    (e) \textbf{Sensor arrangement patterns:}
    To compare different sensory organ arrangements, multiple sensor embedding patterns were implemented inside the biomimetic skin.
    The depth of the accelerometers was varied and used as a reference, around which strain gauge sensors were vertically embedded as densely as possible.
    These sensor arrangement patterns are evaluated in later experiments to investigate how the spatial pattern of sensors affects tactile sensitivity and discrimination performance.
  }
  \label{fig:biomimetic-skin-construction-overview}
\end{figure}

  Human skin is composed of three layers: epidermis, dermis, and subcutaneous tissue, and exhibits a gradient in stiffness, with outer layers being stiffer and inner layers softer.
  However, owing to substantial inter-individual and regional variation, skin stiffness cannot be precisely represented by a single Shore hardness value~\cite{Chatzistergos:2022:ShoreHardness}.
  Based on previous biomimetic sensor studies~\cite{Fishel:2012:BioTac, Hosoda:2006:AnthropomorphicFinger}, we adopted silicone materials for the epidermis and selected a formulation with hardness comparable to that used in BioTac~\cite{Fishel:2012:BioTac}.
  Softer materials were selected for the deeper layers to reproduce the gradual reduction in stiffness.
  The final materials used for each layer are as follows:
  \begin{itemize}
    \item \textbf{Epidermis:} Dragon Skin 30 (silicone rubber, Shore hardness 30A, Smooth-On, Inc.)
    \item \textbf{Dermis:} Ecoflex Gel 2 (silicone gel, Shore hardness 000--34, Smooth-On, Inc.)
    \item \textbf{Subcutaneous tissue:} Soma Foama 15 (foamed silicone sponge, density \SI{240}{\kilogram\per\cubic\meter}, Smooth-On, Inc.)
  \end{itemize}
  Each layer was degassed under vacuum, poured into molds, sequentially cured, and laminated to form the skin structure.
  The thickness of the biological skin varies considerably across individuals and body sites, and reported values also differ among studies.
  For example, the epidermis ranges from several tens to several hundreds of micrometers thick, with the thickest values reported for the plantar surface (\SI{597}{\micro\meter}) \cite{Lintzeri:2022:EpidermalThicknessMetaAnalysis}.
  The dermis ranges from approximately \SI{0.6}{\milli\meter} in the eyelid to \SI{3.0}{\milli\meter} in the skin of the hands and feet \cite{McGrath:2016:GrayAnatomy}, whereas the subcutaneous tissue generally measures approximately \SI{10}{\milli\meter}, although values as high as \SI{36}{\milli\meter} have been reported for abdominal subcutaneous fat \cite{Sim:2014:AppropriatenessInsulinNeedles}.
  Based on these references, we fabricated a biomimetic skin with an epidermis of \SI{0.2}{\milli\meter}, dermis of \SI{2}{\milli\meter}, and subcutaneous tissue of \SI{20.0}{\milli\meter}.
  \figref{fig:biomimetic-skin-construction-overview}a and \figref{fig:biomimetic-skin-construction-overview}b illustrate the correspondence between the layered structure of human skin and that of the fabricated biomimetic skin.
  The overall size of the artificial skin was \SI{90}{\milli\meter} on each side, with sensory receptors embedded in a central \SI{50}{\milli\meter} square region.
  This central region was constructed under three different receptor arrangement patterns (A, B, C), enabling a systematic comparison.
  In human skin, friction ridges have been reported to enhance tactile sensitivity \cite{Scheibert:2009:FingerprintsTactileInformation, Prevost:2009:FingerprintsOrientationSkinVibrations}, and such features have been mimicked in certain biomimetic sensors \cite{Fishel:2012:BioTac}.
  Although the present study primarily focuses on receptor arrangements, we added a ridge-like surface structure to the epidermis (ridge height: \SI{0.2}{\milli\meter}, width: \SI{0.2}{\milli\meter}, spacing: \SI{0.2}{\milli\meter}) to improve clarity of observation and comparison.
  Furthermore, dermal papillae structures \cite{Cauna:1954:NatureFunctionsPapillaryRidges}, which are also believed to enhance tactile sensitivity \cite{Pham:2017:3DMicrostructureDermalPapillae}, were incorporated as ridge–valley patterns with a height of \SI{0.5}{\milli\meter} and width of \SI{1.25}{\milli\meter} at the interface between the epidermis and dermis.

  The cutaneous mechanoreceptors are extremely small and densely distributed \cite{Corniani:2020:TactileInnervationDensities}.
  It is technically infeasible to reproduce all receptor types at the same density as that in human skin because of spatial constraints and current sensor limitations.
  Therefore, in this study we focused on two receptor types that are broadly distributed across both trunk and fingertip skin: Merkel cells, which detect static pressure, and Pacinian corpuscles, which detect vibrations.
  We aimed to enable macroscopic verification by constructing a simplified biomimetic skin incorporating these two receptors.
  For Merkel cells, which are associated with slowly adapting type I (SA1) afferents and respond to sustained pressure, we used strain gauge sensors (General-purpose Foil Strain Gages, KYOWA ELECTRONIC INSTRUMENTS CO., LTD, Tokyo, Japan) whose resistance varies with applied pressure.
  For Pacinian corpuscles, which are associated with rapidly adapting type II (RA2) afferents and respond exclusively to dynamic stimuli, we used MEMS accelerometers (MC3479, Memsic Inc., Shanghai, China).
  By embedding these surrogate receptors within the biomimetic skin, we were able to acquire and analyze sensory signals specific to each arrangement (\figref{fig:biomimetic-skin-construction-overview}c, \figref{fig:biomimetic-skin-construction-overview}d).
  Ruffini endings were excluded because morphological evidence remains limited \cite{Olson:2016:lowthresholdMechanoreceptors, Cobo:2021:HumanDigitalRuffini} and they are thought to primarily encode skin stretch.
  Meissner corpuscles were also excluded, as they are scarce outside the glabrous skin of the fingertips and soles \cite{Wu:2012:MorphologicDiversityCutaneousSensory, Neubarth:2020:MeissnerCorpuscles}, and fine texture discrimination is known to be achievable via Pacinian corpuscles \cite{Bensmaia:2003:VibrationsTexture}.
  Furthermore, the frequency response of the Meissner corpuscles is limited to approximately \SI{50}{\hertz}, which predominantly contributes to slip detection, a range already covered by the accelerometers used here.

  The strain gauges were \SI{5}{\milli\meter} long, \SI{1.4}{\milli\meter} in width, and \SI{14}{\micro\meter} in thickness, and were connected to an Arduino MEGA \cite{Arduino:2015:Arduino,Banzi:2022:GettingStartedArduino} via a dedicated voltage divider circuit.
  The accelerometers consisted of MEMS chips measuring \SI{2}{\milli\meter} $\times$ \SI{2}{\milli\meter} $\times$ \SI{0.92}{\milli\meter}, mounted on \SI{10}{\milli\meter} square circuit boards, which were directly embedded in the skin.
  Sensor data were transmitted from the Arduino MEGA to a PC via serial communication using Robot Operating System (ROS) middleware \cite{Quigley:2009:ROS}.
  The final sampling frequencies were approximately \SI{360}{\hertz} for the strain gauges and \SI{500}{\hertz} for the accelerometers, both of which encompassed the peak response ranges of the corresponding biological receptors.

    In designing the sensory arrangements, the achievable spatial density is constrained by the sensor size and the need to avoid mechanical interference between sensors.
    In this study, a $\SI{20}{\milli\meter} \times \SI{20}{\milli\meter}$ skin block was used as the basic unit, and all sensors were arranged within this area.
    Because Pacinian corpuscles (accelerometers) are larger than Merkel cells (strain gauge sensors), we first defined the placement of the accelerometers and then designed configuration patterns based on their depth within the skin.
    Specifically, three configurations (A, B, and C) were constructed by placing the accelerometer at different depths: near the surface (epidermal side), from the deep dermis to the subcutaneous tissue, and in the deeper subcutaneous region.
    Strain gauge sensors corresponding to Merkel cells were then distributed while avoiding spatial interference with both the accelerometer and other strain gauge sensors.
    Within the $\SI{20}{\milli\meter} \times \SI{20}{\milli\meter}$ skin block, it was determined that up to four strain gauge sensors could be placed per layer without introducing mechanical interference or directional bias, and this constraint was adopted as the basic design principle.
    An overview of the configurations is shown in Fig.~\ref{fig:biomimetic-skin-construction-overview}e.
    Here, “shallow” refers to the epidermal side, and “deep” refers to the subcutaneous side.
    \begin{itemize}
      \item \textbf{A:} The accelerometer is placed near the surface (without being exposed), and four strain gauge sensors are distributed at deeper locations than the accelerometer.
      \item \textbf{B:} The accelerometer is placed from the deep dermis to the subcutaneous tissue, and eight strain gauge sensors are distributed at shallower locations than the accelerometer.
      \item \textbf{C:} The accelerometer is placed in the deeper subcutaneous region, and eight strain gauge sensors are distributed at shallower locations than the accelerometer.
    \end{itemize}
    In configurations B and C, placing eight strain gauge sensors at the same depth leads to spatial interference; therefore, the sensors were distributed across two depth levels.
    This multilayer arrangement allows the spatial density to be increased while maintaining mechanical independence between sensors.
    In configuration B, four sensors were placed at the epidermal–dermal boundary and four in the dermal layer immediately beneath it.
    In configuration C, four sensors were placed in the deep dermis and four at the dermal–subcutaneous boundary (on the subcutaneous side).
    With this design, it is possible to compare how differences in sensory arrangements, varied under physically realizable constraints, affect tactile function.
    In particular, configuration B places strain gauge sensors at the epidermal–dermal boundary and the accelerometer in the deep dermis, and thus is intended to reflect key qualitative characteristics of biological mechanoreceptor organization.

\subsection{Mechanical Characterization by Indentation Testing} \label{subsec:self_organized_skin_construction_mechanical_characterization_methods}

    To evaluate the mechanical properties of the constructed biomimetic skin, indentation tests were performed.
    The test samples were fabricated following essentially the same layered construction described in Section~\ref{subsec:self_organized_skin_construction_methods}, except that the dermal papilla structure was omitted to reduce spatial variability in the indentation response.
    Mechanical characterization was conducted using a tabletop universal testing machine (MCT-2150, A\&D Company, Tokyo, Japan).

    A rigid spherical indenter with a radius of $R = \SI{2.5}{\milli\meter}$ was pressed vertically into the sample surface from a non-contact state at a constant speed of \SI{10.0}{\milli\meter\per\minute}.
    The indenter was displaced and then returned to the initial position, forming a loading-unloading cycle.
    This cycle was repeated ten times, and the force $F$ and displacement $x$ were continuously recorded.

  The contact point $x_c$ between the indenter and the sample was determined as the displacement at which the force first exceeded a threshold defined as the mean plus two standard deviations (mean $+2\sigma$) of the pre-contact region.
  The indentation depth was then defined as $\delta = x - x_c$.

  The Young's modulus of human skin has been reported over a wide range, typically from \SI{10}{\kilo\pascal} to \SI{500}{\kilo\pascal}, depending on the measurement method, model, and experimental conditions \cite{Pailler-Mattei:2008:ElasticMechanicalPropertiesSkin, Abdouni:2017:BiophysicalPropertiesFinger, Pan:1997:InVivoHighFrequencyUltrasound, Boyer:2012:NonContactSkinMechanicalProperties}.
  In some cases, even higher values have been reported \cite{Diridollou:2001:SkinAgeingPhysicalProperties}.
  Therefore, in this study, we estimate the reduced Young's modulus $E^*$ from indentation tests and use it as an approximate indicator to evaluate whether the mechanical softness of the proposed biomimetic skin is comparable to that of human skin.
  Since the thickness of skin varies depending on the measurement location and the adhesive properties of the materials differ from those of biological tissue, precise estimation of the intrinsic Young's modulus is considered difficult.
  Accordingly, the purpose of this analysis is limited to an approximate comparison.

  In static indentation testing, the reduced Young's modulus $E^*$ can be determined from the normal contact stiffness $k_z$, which is defined as the slope of the initial portion of the force-displacement curve.
  Following previous studies \cite{Pailler-Mattei:2008:ElasticMechanicalPropertiesSkin, Abdouni:2017:BiophysicalPropertiesFinger}, the reduced modulus is given by
  \begin{equation}
  E^* = \frac{\sqrt{\pi}}{2}\frac{k_z}{\sqrt{A}},
  \end{equation}
  where $A$ is the projected contact area.

  The tangent stiffness $k_z$ was obtained from the derivative of a third-order polynomial fitted to the loading curve after contact:
  \begin{equation}
  k_z = \frac{dF}{d\delta},
  \end{equation}
  where $\delta$ denotes the indentation depth.

  The contact radius $a$ and projected contact area $A$ were geometrically estimated as
  \begin{equation}
  a = \sqrt{2R\delta - \delta^2}, \quad A = \pi a^2.
  \end{equation}
  For sufficiently small indentation depths ($\delta \ll R$), the Hertzian approximation $a \approx \sqrt{R\delta}$ can be applied.

  However, in this study, a spherical indenter with a radius of \SI{2.5}{\milli\meter} was used, considering the experimental conditions in previous studies \cite{Abdouni:2017:BiophysicalPropertiesFinger}.
  Therefore, the condition $\delta \ll R$ is not strictly satisfied.
  To address this, both the Hertzian approximation and the geometric estimation were used to calculate the reduced Young's modulus $E^*$, and the results were compared.

  When adhesive effects are significant, the contact mechanics should be described using the Johnson–Kendall–Roberts (JKR) theory \cite{Johnson:1971:JKRTheory}.
  However, no significant pull-off force was clearly observed in the present experiments, and therefore adhesive effects were considered to be negligible for the purpose of this analysis.

\subsection{Two-point Discrimination Threshold Task} \label{subsec:self_organized_experiments_hiro_two_point_discrimination_methods}
  To ensure repeatability and reproducibility, we applied indentation stimuli to the skin using HIRO-NX, a dual-arm robot based on NEXTAGE (Kawada Industries Inc., Tokyo, Japan) designed for research use \cite{nextage_robot}.
  Each arm of HIRO-NX has six degrees of freedom and provides high-precision motion, with a positioning repeatability of less than \SI{30}{\micro\meter}.
  The robot is equipped with interchangeable end-effectors, and a six-axis force/torque sensor (IFS-50M31A25-I25, NITTA Corp., Osaka, Japan) was mounted on the wrist.
  Appropriate tools were attached depending on the experimental task.
  The robot was controlled using the open-source Robot Operating System (ROS) \cite{Quigley:2009:ROS}.

\begin{figure}[htbp]
  \centering
  \includegraphics[width=1.0\linewidth]{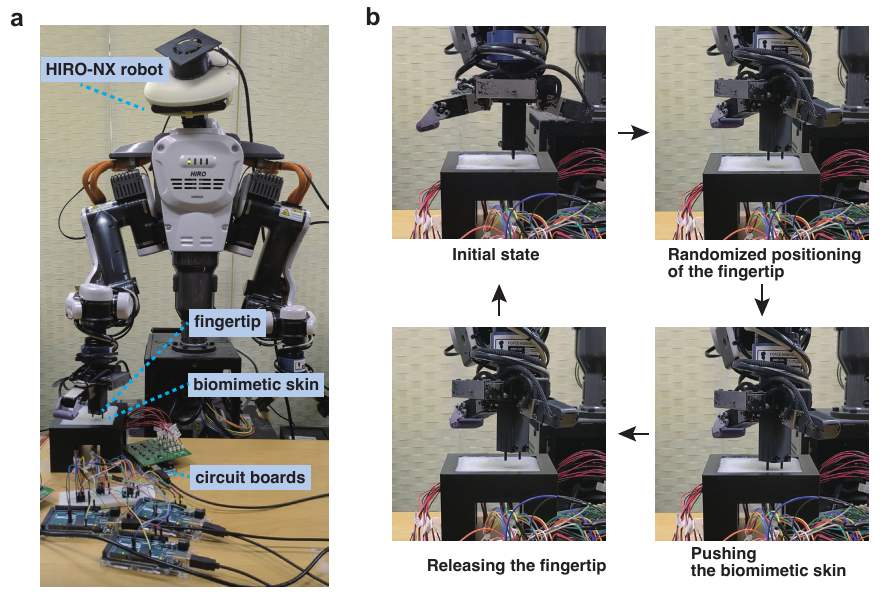}
  \caption{
    Experimental setup for applying point stimuli to biomimetic skin using the HIRO-NX robot.
    (A) Overview of the setup. The HIRO-NX robot's fingertip is equipped with a custom end-effector that delivers one- or two-point tactile stimuli to the biomimetic skin laid on a flat surface.
    The inter-point distance is varied by replacing the end-effector.
    (B) Experimental timeline. The robot applies stimuli with randomized variations in position and orientation, repeating the indentation cycle throughout the experiment.
  }
  \label{fig:hiro-multi-touch-experiment-overview}
\end{figure}

  The experimental setup is illustrated in \figref{fig:hiro-multi-touch-experiment-overview}.
  We prepared four types of end-effectors: one for single-point stimulation and three for two-point stimulation with inter-point distances of \SI{5}{\milli\meter}, \SI{10}{\milli\meter}, and \SI{15}{\milli\meter}.
  For each receptor arrangement pattern, the end-effector was aligned with the center of the pattern, and random noise was added to both the rotation angle (\SIrange{-180}{180}{\degree}) and the position (\SIrange{-10}{10}{\milli\meter}).
  The robot then performed repeated motions of pressing vertically onto the surface of the biomimetic skin and returned to its initial position.
  Random noise of \SIrange{0}{3}{\milli\meter} was also applied to the indentation depth, because pressing to a constant depth could allow discrimination based solely on subtle force differences between single- and two-point contacts.
  For each condition, 100 indentation trials were performed and sensor data were collected.

  To evaluate the performance, the collected sensor data were used in a classification task resembling the two-point discrimination threshold test, where the goal was to distinguish single-touch from multi-touch (two-point) stimuli.
  After removing the initial offset values from each sensor signal, waveform peaks were extracted as features.
  When no peaks were detected, zero padding was applied.
  Using these features, we trained a Random Forest classifier using 200 decision trees.
  Classification performance was evaluated using 5-fold stratified cross-validation, repeated three times with different random seeds, and reported as the mean accuracy and standard deviation across the 15 folds.

\subsection{Small Force Detection Task} \label{subsec:self_organized_experiments_hiro_small_force_detection_methods}
  Human skin can detect even very slight contacts, whereas conventional force sensors typically used in robotics struggle to achieve this sensitivity.
  In this experiment, we quantitatively evaluated the influence of receptor arrangement on the detection of subtle contact stimuli using HIRO-NX.
  First, the reference position was defined as the robot height at which a sheet of paper could no longer pass between the end-effector of HIRO-NX and the biomimetic skin, and this height was recorded as the baseline.
  For each receptor arrangement pattern, the end-effector was aligned with the center of the pattern, and random noise of \SIrange{-10}{10}{\milli\meter} was added to its position.
  The robot then repeated 20 cycles of contact with the skin surface and returned to the initial position, during which sensor data were collected.

  The collected waveform data were preprocessed in the same manner as in the two-point discrimination experiment, and contact was identified by detecting the peaks.
  The number of detected peaks corresponded to the number of successful detections of subtle contacts.
  However, because the skin behaves as a continuum, small responses may appear in sensors even when the stimulation point is not directly above them, but is transmitted through adjacent regions.
  To improve the precision of the contact localization, it is desirable to distinguish such background responses from genuine small-force contacts.
  For this purpose, a threshold was introduced in the peak detection step.
  Specifically, during the two-point discrimination experiment, we statistically evaluated the sensor responses of a given receptor arrangement, whereas another arrangement was being pressed.
  For example, when evaluating small force detection in Pattern C, the sensor responses of Pattern C during the stimulation of Pattern B were used as the reference baseline.
  The mean ($\mu$) and standard deviation ($\sigma$) of these responses are summarized in \tabref{tab:hiro_multi_touch_mutual_stats}.
  In this study, peaks exceeding $\mu + 2\sigma$ were regarded as significant contact detections.

\begin{table}[htbp]
\centering
\begin{tabular}{lcc}
\hline
  & \textbf{Mean} & \textbf{Standard Deviation} \\
\hline
  A (pressed C) & 4.83  & 2.54  \\
  B (pressed C) & 8.02  & 3.55  \\
  C (pressed B) & 36.39 & 16.32 \\
\hline
\end{tabular}
\caption{Mean and standard deviation of sensor responses in non-target patterns.}
\label{tab:hiro_multi_touch_mutual_stats}
\end{table}

\subsection{Passive Tactile Identification Task under Lateral Shear or Stroking Stimuli} \label{subsec:self_organized_experiments_PR2_texture_identification_methods}
  In this experiment, we evaluated how receptor arrangements influence tactile sensing when stroking stimuli were applied to the biomimetic skin described in Section~\ref{subsec:self_organized_skin_construction_methods}.

  To ensure reproducibility, we employed the dual-arm mobile robot PR2~\cite{pr2_robot, Bohren:2011:PR2LessonsLearned}.
  Although preliminary trials were conducted with HIRO-NX, it proved unsuitable for this task owing to its high mechanical stiffness and control gains, which introduced steady-state oscillations that propagated to the sensors.
  In contrast, PR2 is equipped with a counterbalance mechanism that offsets the arm weight, has backdrivable and mechanically compliant joints, and operates with relatively low control gains, making it better suited for stroking experiments.
  Each arm of PR2 has 7 degrees of freedom, allowing precise manipulation.
  In this study, the gripper of one arm was modified to mount test materials with different surface textures, which were then stroked across the biomimetic skin.
  The materials used were metal, cork, roughened cork, and cloth.
  Each was fabricated in a cylindrical shape with a diameter of approximately \SI{30}{\milli\meter}, fixed to the gripper tip to prevent rotation, and gently stroked across the skin.
  The roughened cork was prepared by lathe machining, and the cloth was wrapped around a cork cylinder to maintain the same overall diameter.

\begin{figure}[htbp]
  \centering
  \includegraphics[width=1.0\linewidth]{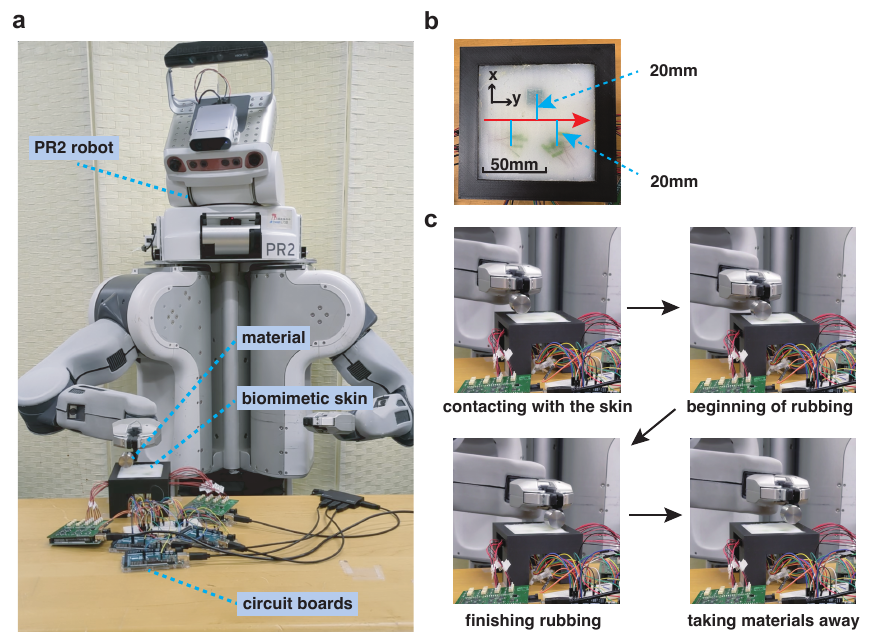}
  \caption{
    Experimental setup for rubbing stimuli with the PR2 robot.
    (a) Overview of the experiment. A material sample is attached to the fingertip of the PR2 robot and rubbed against the biomimetic skin placed on a horizontal plane.
    (b) The material is rubbed along the centerline of the sensory organ arrangement on the biomimetic skin (indicated by the red arrow).
    (c) Time sequence of the experiment. The robot lowers its fingertip, waits 3 seconds after contact, performs the rubbing motion, and then lifts off. This cycle is repeated.
  }
  \label{fig:pr2-rubbing-experiment-overview}
\end{figure}

  The experimental setup is shown in \figref{fig:pr2-rubbing-experiment-overview}a.
  Because it is difficult to precisely define the moment of contact with soft skin, we defined “contact” as the state in which a sheet of paper could no longer pass between the robot gripper and biomimetic skin while the arm was slowly lowered.
  The robot memorized the end-effector height at this contact point, and then performed stroking motions with an additional indentation of \SI{3}{\milli\meter}.
  Although we attempted to ensure precise conditions by using the robot, it was challenging to perfectly align the initial state for each receptor arrangement when performing separate trials.
  Therefore, as shown in \figref{fig:pr2-rubbing-experiment-overview}b, the stroking path was aligned along the midline across the three receptor arrangement patterns to reduce unintended variation between conditions, enabling the simultaneous comparison of all sensors under identical stimuli.
  As depicted in \figref{fig:pr2-rubbing-experiment-overview}c, the end effector was first brought into contact and held until the transient effects from the initial impact dissipated, after which it was stroked across the skin surface and lifted away; this sequence was repeated five times.
  To account for speed-dependent effects, trials were conducted at three different velocities: \SI{2}{\centi\meter\per\second}, \SI{3}{\centi\meter\per\second}, and \SI{6}{\centi\meter\per\second}.

  Sensor data from the accelerometers were processed to extract the frequency-domain features corresponding to frictional interactions with the materials.
  After subtracting the gravitational acceleration from each axis ($x, y, z$), the lateral magnitude is computed as follows:
  \begin{equation}
    \mathrm{acc_{xy}} = \sqrt{\mathrm{acc}_x^2 + \mathrm{acc}_y^2}.
  \end{equation}
  A moving-window RMS (root mean square) was then applied, and segments exceeding a material-specific threshold were identified as stroking intervals.
  For each segment, Fourier transforms were computed, and the following features were extracted across the $x$, $y$, $z$, and $xy$-norm axes:

  \begin{itemize}
    \item \textbf{RMS (Root Mean Square)}: overall signal energy
    \begin{equation}
      \mathrm{RMS} = \sqrt{\frac{1}{N}\sum_{n=0}^{N-1} x[n]^2}.
    \end{equation}

    \item \textbf{Dominant Frequency}: frequency with maximum spectral energy
    \begin{equation}
      \mathrm{Dominant\ Frequency} = \underset{f \geq f_{\min}}{\argmax}\ |X(f)|^2, \quad f_{\min} = 2\ \mathrm{Hz}.
    \end{equation}

    \item \textbf{Spectral Centroid}: weighted average of the spectrum
    \begin{equation}
      f_{\mathrm{centroid}} = \frac{\sum_{f} f \cdot |X(f)|^2}{\sum_{f} |X(f)|^2}.
    \end{equation}

    \item \textbf{Spectral Bandwidth}: spread around the dominant frequency
    \begin{equation}
      \mathrm{Bandwidth} = \sqrt{\frac{\sum_{f}|X(f)|^2 \cdot (f - f_{\mathrm{dom}})^2}{\sum_{f}|X(f)|^2}}.
    \end{equation}

    \item \textbf{Zero-Crossing Rate (ZCR)}: normalized count of sign changes
    \begin{equation}
      \mathrm{ZCR} = \frac{1}{T}\sum_{n=1}^{N-1}\mathbb{I}[x[n-1]\cdot x[n] < 0].
    \end{equation}
  \end{itemize}

  Using these features, we trained Random Forest classifiers to discriminate among the four materials.
  The performance was evaluated by 5-fold cross-validation.
  For each receptor arrangement pattern (A, B, and C), data from five repetitions per material were used.
  Classifiers were trained separately for each stroking speed (\SI{2}{\centi\meter\per\second}, \SI{3}{\centi\meter\per\second}, \SI{6}{\centi\meter\per\second}).

\FloatBarrier

\section{Results} \label{sec:self_organized_skin_results}

    This section first describes the evaluation of the mechanical properties of the constructed biomimetic skin using indentation testing, followed by the results of three discrimination tasks: two-point discrimination, small-force detection, and texture identification.
  
  For each task, the recognition accuracy was quantitatively evaluated for the three receptor arrangement patterns (A, B, and C), and the influence of receptor depth and density on performance was compared.

\subsection{Results of Mechanical Characterization by Indentation Testing} \label{subsec:self_organized_indentation_material_test_results}

\begin{figure}[htbp]
  \centering
  \includegraphics[width=1.0\linewidth]{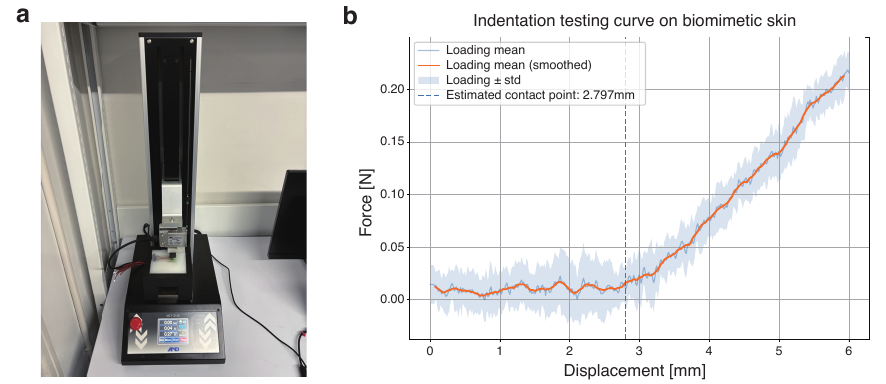}
  \caption{
    (a) Experimental setup of the indentation test using the tabletop universal testing machine.
    (b) Averaged force–displacement curve obtained from 10 loading–unloading cycles. The shaded region represents the standard deviation. The dashed line indicates the estimated contact point ($x_c = \SI{2.797}{\milli\meter}$).
  }
  \label{fig:indentation-material-test-results}
\end{figure}

  \figref{fig:indentation-material-test-results} shows the averaged force-displacement curve obtained from 10 loading-unloading cycles, along with the standard deviation.
  No explicit pull-off force was observed.
  A clear hysteresis was observed between the loading and unloading curves, suggesting viscoelastic behavior of the sample.
  The contact point was estimated as $x_c = \SI{2.797}{\milli\meter}$.
  Based on this, the indentation depth was defined as $\delta = x - x_c$.

  The reduced Young's modulus $E^*$ was calculated at representative indentation depths of $\delta = \SI{1.0}{\milli\meter}$, \SI{1.5}{\milli\meter}, \SI{2.0}{\milli\meter}, and \SI{2.5}{\milli\meter}, using the contact area $A$ and the tangent stiffness $k_z$.
  The tangent stiffness $k_z$ was obtained as the derivative of a third-order polynomial fitted to the loading curve after contact.
  The contact area $A$ was estimated based on both the geometric formulation and the Hertzian contact model.
  Using the geometric approximation, the reduced Young's modulus was estimated as \SI{15.21 \pm 0.14}{\kilo\pascal}, whereas the Hertzian approximation yielded \SI{17.29 \pm 1.89}{\kilo\pascal}.

\subsection{Results of Two-point Discrimination Threshold Task} \label{subsec:self_organized_experiments_hiro_two_point_discrimination_results}
\begin{figure}[htbp]
  \centering
  \includegraphics[width=\linewidth]{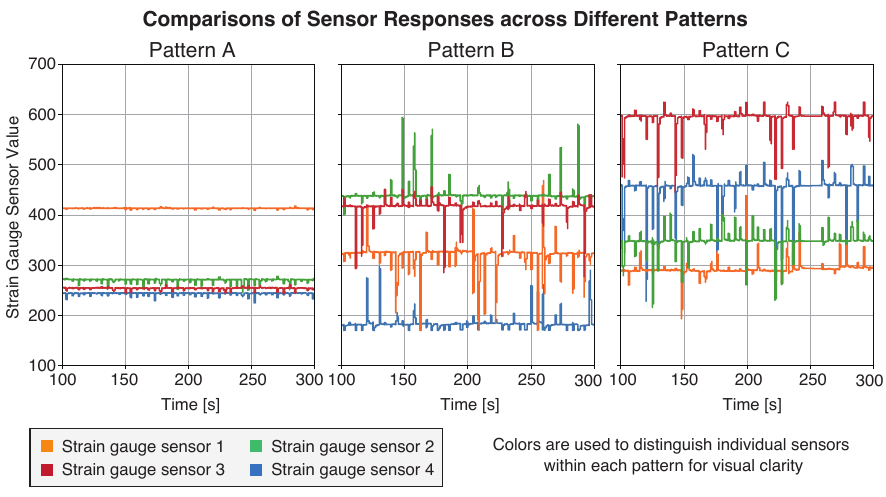}
  \caption{
      Comparison of strain gauge sensor responses to deformation in the biomimetic elastic skin.
    The figure presents representative sensor outputs from embedded strain gauges under a two-point indentation with a \SI{15}{mm} inter-point distance.
    Three sensor embedding patterns (Pattern A, B, and C) are compared.
    For each pattern, responses from four surface-level sensors are plotted to assess how sensor placement affects sensitivity to localized pressure.
    The plots show a selected time segment from the experimental recordings.
      Different colors are used to distinguish individual sensors within each pattern for visual clarity and do not imply correspondence across patterns or differences in sensor function.
  }
  \label{fig:hiro-multi-touch-experiment-data}
\end{figure}

  An example of recorded waveform data is presented in \figref{fig:hiro-multi-touch-experiment-data}.
  Visual inspection indicates that in Pattern A, where the strain gauges were placed in deeper layers, the sensitivity was lower and the responses were weaker compared to other configurations.

  For each configuration (A, B, C) and each condition (single-touch and two-point stimuli at \SI{5}{mm}, \SI{10}{mm}, and \SI{15}{mm}), the mean absolute change in sensor response per trial was computed.
  Descriptive statistics are summarized in \tabref{tab:hiro_multi_touch_strain_stats}.
  The ANOVA revealed a significant main effect of condition ($F(2,1179)=\num{386.93},\; p<\num{1e-129},\; \eta^{2}=\num{0.396}$).
  Post-hoc Tukey HSD tests confirmed significant differences for all pairs (A vs B, A vs C, and B vs C; $p<.001$).
  Welch's $t$-tests further supported these results: A vs B ($t=-28.60,\; p<.001,\; g=-2.03$), A vs C ($t=-22.40,\; p<.001,\; g=-1.61$), B vs C ($t=7.80,\; p<.001,\; g=0.56$).
  The effect sizes ranged from medium to large, indicating robust differences.
  Additionally, the standard deviations of the raw sensor responses (without trial averaging) were A: $\num{7.35}$, B: $\num{36.01}$, C: $\num{33.58}$.
  Taken together, these results suggest that strain gauges embedded closer to the surface produced stronger but more variable responses, whereas a deeper placement yielded weaker but more stable responses.

\begin{figure}[htbp]
  \centering
  \includegraphics[width=1.0\linewidth]{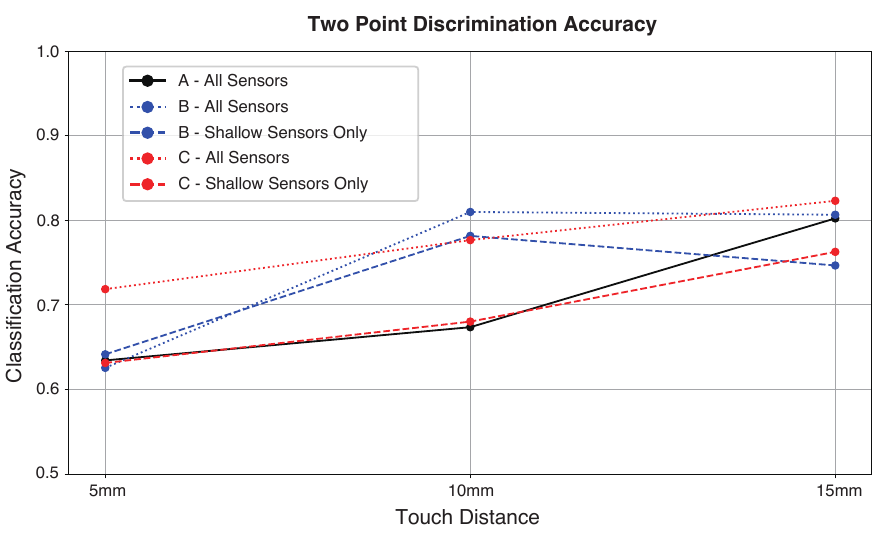}
  \caption{
    Classification accuracy for single- vs.\ multi-touch discrimination at varying inter-point distances.
    This plot illustrates the classification performance of a machine learning model distinguishing single-point and two-point stimuli at distances of \SIrange{5}{15}{\milli\meter}, simulating a two-point discrimination threshold test.
    Curves correspond to sensor arrangement patterns: black (A), blue (B), red (C); line styles indicate sensor usage—solid: sensors in A; dashed: only shallow sensors in B/C; dotted: all sensors in B/C.
  }
  \label{fig:hiro-two-point-discrimination-accuracy}
\end{figure}

  The classification performance was evaluated using datasets combining single-touch and multi-touch (two-point) stimuli at varying distances.
  The results are shown in \figref{fig:hiro-two-point-discrimination-accuracy}.
  Error bars are omitted for clarity, but are provided in the Supplementary Information (\figref{fig:hiro-two-point-discrimination-accuracy-with-error-bar-A-all}, \figref{fig:hiro-two-point-discrimination-accuracy-with-error-bar-B-all}, \figref{fig:hiro-two-point-discrimination-accuracy-with-error-bar-B-shallow}, \figref{fig:hiro-two-point-discrimination-accuracy-with-error-bar-C-all}, \figref{fig:hiro-two-point-discrimination-accuracy-with-error-bar-C-shallow}).
  Owing to physical constraints, Pattern A included only four sensors, whereas Patterns B and C allowed for evaluation with either four shallow sensors only (Shallow Sensors Only) or all eight sensors (All Sensors).
  Across all configurations, the classification accuracy improved as the inter-point distance increased, which is consistent with classical findings from human two-point discrimination tests.
  Moreover, in both Patterns B and C, the performance was higher when using all sensors compared to shallow sensors only, highlighting the contribution of sensor number and density—consistent with the physiological distribution of Merkel cells at high density near the epidermal–dermal boundary.

  A closer comparison of Patterns A, and B (Shallow Sensors Only), and C (Shallow Sensors Only) revealed that, at \SI{5}{\milli\meter}, the classification accuracy was modest (low 60\% range) across all patterns.
  At \SI{10}{\milli\meter}, Pattern B exhibited a sharper improvement in accuracy than Patterns A and C.
  In contrast, at \SI{15}{\milli\meter}, the accuracy of Patterns A and C improved more markedly, whereas Pattern B plateaued.
  This difference can be understood in terms of the receptive field size, in which strain gauges were positioned most superficially, resulting in narrower receptive fields, whereas those in A and C were embedded deeper, corresponding to broader receptive fields.
  As a result, Pattern B was particularly effective at a closer spacing of \SI{10}{\milli\meter}, and its narrow receptive fields allowed finer spatial discrimination.
  However, at the wider spacing of \SI{15}{\milli\meter}, random positional noise meant that some stimuli extended beyond the \SI{20}{\milli\meter}-square sensor region, and the limited receptive fields of Pattern B failed to capture these stimuli effectively.
  In contrast, the broader receptive fields of Patterns A and C enabled them to detect such stimuli, leading to improved classification accuracy at \SI{15}{\milli\meter}.
  These findings suggest that superficial sensor placement is advantageous for fine spatial resolution, in agreement with the shallow distribution of Merkel cells near the epidermal–dermal junction in human skin.

  On the other hand, Pattern C (All Sensors) also achieved high accuracies overall, and in some cases surpassed Pattern B, particularly at \SI{5}{\milli\meter} and \SI{15}{\milli\meter}.
  One likely explanation is that the limited number of embedded sensors in this study introduced variability, which was most pronounced for the superficially placed sensors in Pattern B, thereby reducing the relative discriminative performance.
  At \SI{5}{\milli\meter}, this variability caused the accuracy distributions of Pattern B and Pattern C (All Sensors) to overlap considerably when the standard deviations were taken into account.
  At \SI{15}{\milli\meter}, the narrow receptive fields of Pattern B limited the improvement, whereas Pattern C benefited from deeper placement and a greater number of sensors.
  Future improvements in the sensor density and miniaturization could reduce this variability.
  Nevertheless, even under these constraints, the pronounced accuracy gain of Pattern B at the short spacing of \SI{10}{\milli\meter}, where it outperformed the other patterns, suggests a potential functional benefit of human-like receptor arrangements in fine-scale discrimination.

\begin{table}[htbp]
\centering
\begin{tabular}{lccccccc}
\hline
  {Condition} & {Sample size} & {Mean} & {SD} & {Median} & {Q1} & {Q3} \\
\hline
  A & 397 & 7.298464  & 3.272834  & 7.195545  & 5.230198  & 9.537129 \\
  B & 397 & 29.237356 & 14.931334 & 28.960396 & 17.573020 & 38.785891 \\
  C & 388 & 21.651800 & 12.197818 & 20.090965 & 12.208540 & 29.542698 \\
\hline
\end{tabular}
\caption{Mean and standard deviation of strain gauge responses for the three sensor arrangement patterns (A, B, and C).}
\label{tab:hiro_multi_touch_strain_stats}
\end{table}

\subsection{Results of Small Force Detection Task} \label{subsec:self_organized_experiments_hiro_small_force_detection_results}
  The number of detections in response to the weak contact stimuli is summarized in \tabref{tab:hiro_light_touch_detection_rate}.
  For each receptor pattern (Patterns A, B, and C), the number of events detected when the peak detection threshold was set to $\mu + 2\sigma$.
  The results revealed a clear trend: the detection performance was highest for Pattern B, followed by Pattern C, and then Pattern A.
  In other words, the ability to detect small forces increased when the strain gauge sensors (representing Merkel cell–like receptors) were placed closer to the superficial layer of the skin.
  This finding suggests that superficial placement of receptors enhances sensitivity to subtle contacts, consistent with the biological fact that Merkel cells are distributed near the epidermal–dermal junction in human skin.

\begin{table}[htbp]
\centering
\begin{tabular}{lc}
\hline
\textbf{Pattern} & \textbf{Detection Count (count/20)} \\
\hline
  A & 0  / 20 \\
  B & 17 / 20 \\
  C & 3  / 20 \\
\hline
\end{tabular}
\caption{Detection rates of very light touch stimuli for each sensor arrangement pattern. The values indicate the percentage of correct detections out of 20 trials.}
\label{tab:hiro_light_touch_detection_rate}
\end{table}

\subsection{Results of Passive Tactile Identification Task under Lateral Shear or Stroking Stimuli} \label{subsec:self_organized_experiments_PR2_texture_identification_results}
\begin{figure}[htbp]
  \centering
  \includegraphics[width=1.0\linewidth]{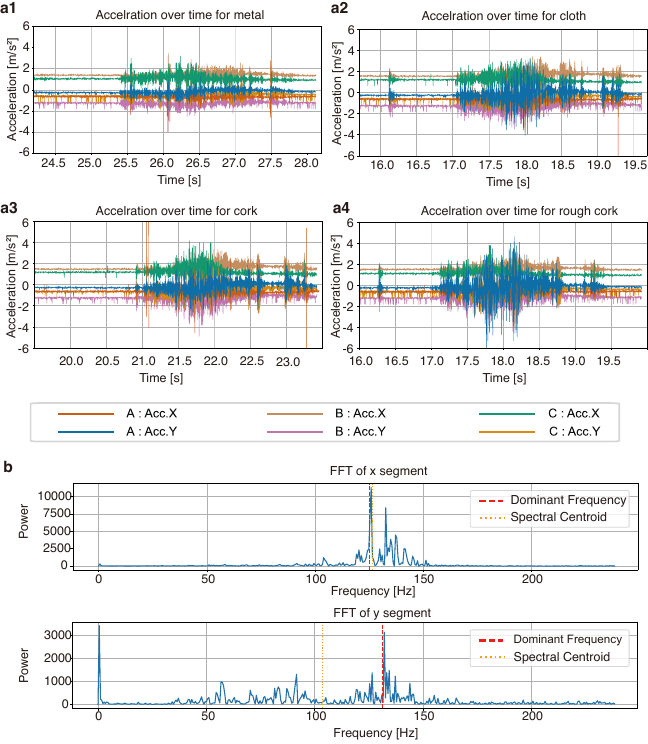}
  \caption{
    Sensor responses and spectral analysis during rubbing experiments.
    (a) Comparison of acceleration sensor waveforms for four materials (a1:metal, a2:cloth, a3:cork, a4:rough cork) rubbed against the biomimetic skin using the robot fingertip. The data shown here corresponds to a rubbing speed of \SI{3}{\centi\meter\per\second}.
    (b) Example of frequency-domain analysis using Fourier transform. The plot shows the FFT result for the cloth material at a speed of \SI{3}{\centi\meter\per\second}.
  }
  \label{fig:pr2-rubbing-experiment-detail}
\end{figure}

  An example of the accelerometer waveforms obtained during stroking is shown in \figref{fig:pr2-rubbing-experiment-detail}a.
  Each trace corresponds to one material stroked at \SI{3}{\centi\meter\per\second}, and clear differences in the sensor responses can be observed across the materials.
  Fourier transforms were performed from the extracted waveform segments, as illustrated in \figref{fig:pr2-rubbing-experiment-detail}b, and features such as the RMS and spectral centroid were calculated.

  The mean classification accuracies obtained using the Random Forest classifier are summarized in \tabref{tab:pr2-texture-randomforest-accuracy}.
  While Pattern B consistently achieved the highest overall accuracy across all stroking speeds, a high classification performance was observed for all receptor arrangement patterns.

\begin{figure}[htbp]
  \centering
  \includegraphics[width=1.0\linewidth]{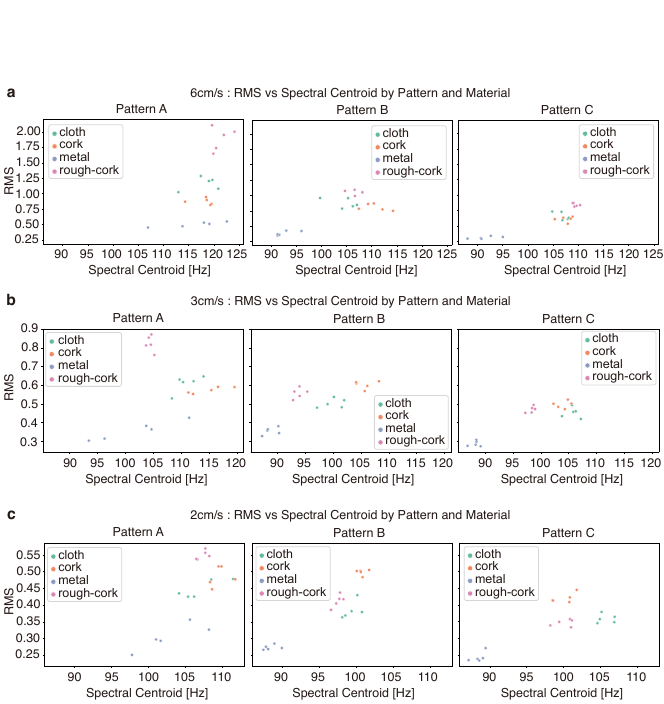}
  \caption{
    Visualization of biomimetic skin responses to four materials across three sensory organ arrangements at different rubbing speeds.
    This figure shows the distribution of tactile sensor responses when four materials (cloth, cork, metal, and rough cork) are rubbed against the biomimetic skin by the PR2 robot at three different speeds: (a) 6 cm/s, (b) 3 cm/s, and (c) 2 cm/s.
    For each condition, spectral centroid (x-axis) and RMS amplitude (y-axis) are plotted to represent the frequency and intensity characteristics of the responses.
    Data from all three sensory organ arrangements (Pattern A, B, and C) are included, illustrating how both material type and sensor layout affect the tactile response under varying dynamic conditions.
  }
  \label{fig:pr2-rubbing-experiment-scattering-plot}
\end{figure}

  A previous work \cite{Fishel:2012:BioTacBayesian} suggested that RMS is associated with the roughness of a texture, whereas the spectral centroid is related to its fineness.
  To visualize these relationships, the distributions of the four materials were plotted in two-dimensional space defined by the RMS and spectral centroid, as shown in \figref{fig:pr2-rubbing-experiment-scattering-plot}.
  The scatter plots display the results for stroking at \SI{6}{\centi\meter\per\second}, \SI{3}{\centi\meter\per\second}, and \SI{2}{\centi\meter\per\second}, using the weighted average spectral centroid along the $x$ and $y$ axes.
  The plots reveal that Pattern A, where vibration-sensitive receptors were embedded more superficially, exhibited larger RMS differences among materials as well as greater dispersion of spectral centroid values.
  This observation was further quantified by calculating the trace of the two-dimensional covariance matrix composed of the RMS and spectral centroid values (\tabref{tab:cov_trace_top20}), with the top \SI{20}{\percent} highlighted in red.
  The statistics confirmed the tendency for greater variance in Pattern A.

  Through these stroking experiments with PR2, we quantitatively evaluated how receptor arrangements in biomimetic skin influence tactile identification.
  All three receptor patterns achieved high classification performance, but Pattern B—closest to the human configuration—provided the most stable and accurate discrimination across conditions.
  
    In addition, placing vibration-sensitive sensors near the skin surface, as in Pattern A, tended to increase the dispersion of the spectral centroid, which may influence the stability of discrimination.
    At the same time, however, the larger RMS differences among materials indicate the potential to capture richer vibrational information.

\begin{table}[ht]
\centering
\caption{Accuracy for each pattern across different speeds}
\begin{tabular}{|c|c|c|c|}
\hline
\textbf{Speed} & \textbf{Pattern A} & \textbf{Pattern B} & \textbf{Pattern C} \\
\hline
\SI{2}{\centi\meter\per\second} & 0.900 & 0.950 & 0.950 \\
\hline
\SI{3}{\centi\meter\per\second} & 0.900 & 0.950 & 0.750 \\
\hline
\SI{6}{\centi\meter\per\second} & 0.900 & 0.900 & 1.000 \\
\hline
\end{tabular}
\label{tab:pr2-texture-randomforest-accuracy}
\end{table}

\begin{table}[htbp]
\centering
\caption{Covariance Trace by Speed, Pattern, and Material (Top 20\% Highlighted)}
\begin{tabular}{|c|c|c|c|c|c|}
\hline
\textbf{Pattern} & \textbf{Speed} & \textbf{metal} & \textbf{cloth} & \textbf{cork} & \textbf{rough-cork} \\
\hline
  \multirow{3}{*}{A} & 1000 & \textbf{36.041} & \textbf{9.493} & 4.589 & 3.432 \\
   & 2000 & \textbf{51.41} & 4.955 & \textbf{11.364} & 0.314 \\
   & 3000 & \textbf{16.843} & \textbf{8.494} & 1.842 & 0.585 \\
\hline
  \multirow{3}{*}{B} & 1000 & 4.091 & \textbf{8.238} & 6.822 & 2.046 \\
   & 2000 & 1.796 & 3.893 & 2.83 & 1.006 \\
   & 3000 & 1.012 & 1.233 & 0.383 & 0.404 \\
\hline
  \multirow{3}{*}{C} & 1000 & 6.712 & 1.705 & 1.991 & 0.347 \\
   & 2000 & 0.731 & 1.529 & 1.695 & 0.448 \\
   & 3000 & 0.783 & 1.272 & 2.213 & 1.652 \\
\hline
\end{tabular}
\label{tab:cov_trace_top20}
\end{table}

\FloatBarrier

\section{Discussion} \label{sec:self_organized_skin_discussion}
  In this study, we constructed a biomimetic skin using robotic technologies and designed models in which groups of sensors mimicking mechanoreceptors were embedded in different spatial patterns.
  This framework allowed us to quantitatively evaluate how variations in receptor arrangement influence the tactile function.

    Several limitations associated with the physical implementation of the biomimetic skin should be acknowledged.
    First, although the materials were selected based on prior work and a layered biomimetic skin structure was fabricated, it is inherently difficult to fully reproduce the complex properties of human skin, which consists of multiple heterogeneous tissues.
    Nevertheless, the reduced Young's modulus $E^*$ estimated from the indentation material test was approximately \SIrange{15}{17}{\kilo\pascal}.
    Although reported values for human skin vary depending on the measurement method and anatomical location, typically ranging from approximately \SI{10}{\kilo\pascal} to several hundred kilopascals,
    values obtained from contact indentation and related non-contact methods relevant to the indentation-based evaluation used in this study are often reported in the range of approximately \SI{10}{\kilo\pascal} to around \SI{100}{\kilo\pascal} \cite{Pailler-Mattei:2008:ElasticMechanicalPropertiesSkin,Abdouni:2017:BiophysicalPropertiesFinger}.
    In this context, the reduced Young's modulus estimated in this study can be regarded as being of a comparable order of magnitude.
    The estimation of $E^*$ in this study is based on several simplifying assumptions, including the neglect of adhesion effects and the use of approximate expressions for the contact area.
    However, the consistency between the results obtained using geometric approximation and Hertzian approximation suggests that the estimated values are reasonably reliable.
    Therefore, the indentation test results support the validity of the proposed biomimetic skin as a physically plausible experimental platform for investigating tactile sensing mechanisms.
    It should also be noted that the proposed biomimetic skin has a simplified layered structure and material composition, and thus differs from actual biological tissues in several aspects.
    Accordingly, the estimated reduced Young's modulus should be interpreted as an effective or apparent mechanical property rather than an intrinsic material constant.

    Second, there are also limitations in the mimicking of mechanoreceptors.
    The strain gauge sensors and accelerometers used in this study are on the millimeter scale, which is substantially larger and less densely distributed than their biological counterparts such as Merkel cells (diameter $\sim$\SI{10}{\micro\meter}) and Pacinian corpuscles (diameter $\sim$\SI{1}{\milli\meter}).
    It is well known that tactile receptor density varies significantly depending on body region and receptor type.
    Here, we refer to afferent fiber density, as a single fiber can innervate one or multiple mechanoreceptors, providing a more consistent measure of sensory distribution.
    For example, slowly adapting type I (SAI) afferent fibers, which are associated with Merkel cells, are densely distributed in the fingertip at approximately \SI{70}{\per\centi\meter\squared}, whereas their density decreases to approximately \SI{10}{\per\centi\meter\squared} in the palm.
    In contrast, fast-adapting type II (FAII) afferent fibers, which are associated with Pacinian corpuscles, are distributed at relatively low densities across the hand, with approximately \SI{10}{\per\centi\meter\squared}, and up to approximately \SI{25}{\per\centi\meter\squared} in the fingertip \cite{Corniani:2020:TactileInnervationDensities}.
    In comparison, the sensor density in this study is substantially lower than that of biological mechanoreceptors due to limitations in sensor size and implementation.
    Such structural and dimensional differences may lead to differences in responsiveness to fine physical phenomena.
    Therefore, the results of this study should not be interpreted as a quantitative reproduction of biological sensory receptor arrangements, but rather as a comparative evaluation of different spatial configurations under physical constraints.
    The sensory arrangements examined in this study were designed to reflect the qualitative spatial relationships of mechanoreceptors within the layered structure of the skin.
    In biological tissue, Merkel cell complexes are located in the basal layer of the epidermis, whereas Pacinian corpuscles are distributed in the deep dermis and subcutaneous tissue.
    The biologically motivated configuration in this study (Pattern B) was designed to approximate this general depth relationship as closely as possible.
    In contrast, the counterfactual configurations (Patterns A and C), which are difficult to examine in biological systems, were constructed by modifying this depth relationship within physically realizable constraints.
    Future work should incorporate higher-density and more finely scaled sensing elements, as well as systematically vary spatial parameters, to further investigate the robustness and generality of the trends observed in this study.

    In addition to these differences in scale and density, the mechanical properties of the sensor elements themselves also differ from those of biological receptors.
    In particular, the MEMS accelerometers used in this study have relatively rigid package structures, whereas biological Pacinian corpuscles consist of compliant, fluid-filled lamellar structures.
    Embedding such rigid inclusions within a soft elastic medium can alter the local mechanical field, potentially introducing effects such as stress concentration and partial shielding of deformation transmission.
    These effects may be especially pronounced when the accelerometer is placed near the surface, as in Pattern A, where it may influence how mechanical stimuli propagate to deeper regions.
    However, in the results of the small-force detection task, the detection performance was already reduced in Pattern C—where the strain gauge sensors corresponding to Merkel cells are located deeper than in Pattern B—compared to Pattern B.
    This result indicates that the depth of the receptors is one of the dominant factors affecting tactile sensitivity.
    From this perspective, a similar tendency may also arise in Pattern A, where the strain gauge sensors corresponding to Merkel cells are positioned even deeper.
    Therefore, although the rigidity of the accelerometer may contribute to local mechanical perturbations, these tendencies can also be explained by differences in receptor arrangement.
    As such, the comparative discussion based on differences in sensor arrangement is not considered to be substantially undermined.
    In future work, the introduction of mechanically compliant sensors, such as soft or gel-based vibration sensors, with mechanical impedance closer to that of biological receptors is expected to further enhance the discussion of biomimetic tactile systems.

  Considering these limitations, we now turn to the interpretation of our findings.
  For the detection of small forces, the configuration with pressure-sensitive sensors embedded in the superficial layer (Pattern B) demonstrated the highest performance.
  In contrast, configurations with deeper sensors exhibited broader receptive fields and were more prone to cross-activation by neighboring stimuli, making them less suitable for the local discrimination of subtle forces.
  A similar trend was observed in the two-point discrimination task, in which superficial placement facilitated the separation of closely spaced contact points.
  This effect can be attributed to the narrower receptive fields of shallow sensors, which accentuate the differences between the single- and two-point stimuli in the sensor responses.
  Furthermore, a higher sensor density also contributed to improved discriminability, a result consistent with the high-density distribution of Merkel cells near the epidermal–dermal boundary in human skin.
  These observations support the view that the structural features of biological receptor organization critically shape tactile performance.
  In the texture identification task under stroking stimuli, the most stable and accurate discrimination was achieved with a receptor arrangement resembling the human condition (Pattern B).
  Nevertheless, high levels of accuracy were attained across all tested configurations, indicating that texture cues are relatively robust to variations in the sensor placement.
  Interestingly, when vibration-sensitive sensors were positioned closest to the surface (Pattern A), the spectral centroid exhibited greater variance, whereas the RMS differences across materials became more pronounced.
  The larger variance of the spectral centroid suggests that superficial placement reduces the filtering effect of the overlying tissue, allowing a broader range of vibrational information to be transmitted.
  At the same time, stronger RMS differentiation indicates that superficial vibration sensors may provide heightened sensitivity to subtle material-dependent differences.

    To facilitate a cross-task comparison among different tactile tasks (texture classification, small-force detection, and two-point discrimination), we introduce the Tactile Discrimination Performance Index (TDPI), as defined in \equref{eq:tdpi_en}, as an auxiliary metric.
    In this study, to account for differences in scale and task difficulty, the performance of each task was first normalized based on its chance level, followed by standardization across patterns (z-score normalization).
    For texture classification and two-point discrimination, the chance levels were determined based on the number of classes, while for small-force detection, the detection rate itself was used as the performance metric.
    For the two-point discrimination task, the performance was defined as the average classification accuracy across three stimulus distances (\SI{5}{\milli\meter}, \SI{10}{\milli\meter}, and \SI{15}{\milli\meter}).
    To account for variability across experimental conditions, a penalty term based on uncertainty was introduced.
    In this study, since the variability across sliding speeds in the texture classification task can be physically interpreted in terms of differences in friction and vibration transmission,
    the uncertainty was defined as the standard deviation of classification accuracy across different speeds and applied only to the texture classification task.
    The TDPI is defined as follows:
    \begin{equation}
      \begin{aligned}
        \mathrm{TDPI}_i &= \sum_k w_k \left( z_{ik} - \lambda \tilde{u}_{ik} \right) \\
        1 &= \sum_k w_k
        \end{aligned}
      \label{eq:tdpi_en}
    \end{equation}
    Here, (i) denotes the sensor arrangement pattern (Pattern A, B, C), and (k) denotes the tactile task (texture classification, small-force detection, and two-point discrimination).
    The term $z_{ik}$ represents the normalized performance of pattern $i$ for task $k$, and $\tilde{u}_{ik}$ denotes the normalized uncertainty.
    This index is not intended to provide a rigorous scalar evaluation of performance across different tactile tasks, but rather to offer an intuitive summary of performance trends across multiple tasks.

\begin{figure}[htbp]
    \centering
    \includegraphics[width=1.0\linewidth]{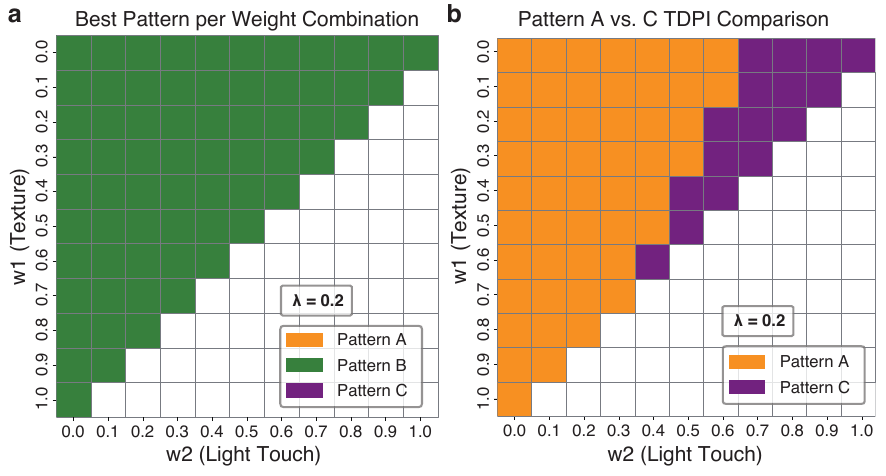}
    \caption{
      TDPI-based selection of the best sensory pattern under varying task weight combinations.
      \textbf{(a)} Heatmap showing the best-performing pattern (A, B, or C) based on the Tactile Discrimination Performance Index (TDPI) when all three patterns are compared.
      \textbf{(b)} Heatmap comparing only patterns A and C.
      The axes represent task weights: $w_1$ (vertical) for texture classification, $w_2$ (horizontal) for light touch detection, and $w_3 = 1 - w_1 - w_2$ for two-point discrimination.
      The variability penalty coefficient for texture classification is fixed at $\lambda = 0.2$ in both panels.
    }
    \label{fig:tdpi-comparison}
  \end{figure}

    For the comparison including all three patterns (A, B, and C), \figref{fig:tdpi-comparison}a shows the sensor arrangement that achieves the highest TDPI score when the weights are varied with a fixed value of $\lambda = 0.2$.
    We additionally confirmed that, even when $\lambda$ is varied, Pattern B consistently exhibits relatively high TDPI scores over a wide range of weight combinations.
    To further examine task-dependent differences, we compare Pattern A and Pattern C by excluding Pattern B, as shown in \figref{fig:tdpi-comparison}b.
    The results obtained by varying $\lambda$ are provided in the Supplementary Material (\figref{fig:ac-tdpi-comparison-by-lambda}).
    When the contribution of small-force detection is relatively reduced and texture recognition is emphasized, Pattern A becomes favorable in certain regions.
    In contrast, when small-force detection is more strongly weighted, Pattern C outperforms Pattern A.
    The comparison between Pattern A and Pattern C indicates that the preferable sensor arrangement may vary depending on which tactile functions are emphasized.
    In this context, the observation that Pattern B, motivated by biological sensor arrangements, exhibits high performance across the passive tactile tasks considered in this study suggests that biologically motivated sensor configurations may be advantageous for certain tactile functions.
    However, since this study is based on comparisons using specific physical configurations, the results should be interpreted as indicative trends within the limited design space explored.

    From a robotics perspective, the results suggest that multimodal tactile sensing capable of distinguishing pressure and texture can be achieved without relying on complex structures or advanced measurement systems, such as those described in Section \ref{sec:self_organized_skin_introduction}.
    Instead, embedding sensors within an elastic medium using biomimetic arrangements may be sufficient.
    This provides a design insight into how sensory arrangement itself contributes to tactile function.
    Furthermore, the results can also be interpreted as an example showing that sensory arrangement may influence perceptual performance through interactions between body structure and the environment.
    This perspective may offer a complementary viewpoint for examining the relationship between body configuration and perception–action processes.

    In this study, we conducted experiments on tactile tasks based on human behavioral patterns, such as pressing and stroking.
    However, when focusing solely on vibratory stimuli, configurations in which vibration-sensitive sensors were placed near the surface suggested the potential to capture richer information.
    This finding implies that, in species with different behavioral patterns, sensory arrangements that place greater emphasis on vibration detection may be functionally advantageous.
    Indeed, examples of relatively superficial distributions of vibration-sensitive receptors have been reported in biological systems.
    For instance, Eimer's organs in the nose of moles \cite{Marasco:2006:EimersOrgan, Catania:2011:StarNosedMoleTouch} consist of layered epidermal structures, beneath which Pacinian-like corpuscles are located in the superficial dermis.
    Such structures can be interpreted as an example of an arrangement suited for detecting vibratory stimuli.
    Similarly, in birds, particularly in waterfowl of the family Anatidae such as mallards, vibration-related mechanoreceptors, including Grandry and Herbst corpuscles,
    are known to be densely distributed around the beak \cite{Berkhoudt:1979:HerbstGrandryCorpusclesMallard, Avilova:2018:LamellirostralBirdsMechanoreceptorOrgans, Ziolkowski:2022:BirdsTactileSensation}.
    These receptors have also been reported to be predominantly distributed in the superficial dermis.
    In particular, Herbst corpuscles are interpreted as mechanoreceptors corresponding to human Pacinian corpuscles, and they exhibit sensitivity to high-frequency vibrations and relatively small size.
    Detailed studies in ostriches \cite{Crole:2014:HerbstCorpusclesOropharynx} and ducks \cite{Soliman:2017:BeakSensoryUnits} have investigated the distribution and proportion of Herbst corpuscles.
    Notably, certain regions of the duck beak have been reported to contain a higher density of Herbst corpuscles than Merkel cells, which are primarily responsible for pressure sensing.
    Such distributions have been interpreted as reflecting an arrangement that places greater emphasis on vibration detection, associated with behavioral strategies that rely on detecting subtle vibrations to locate prey in muddy or aquatic environments.

    Mechanoreceptors have developed and changed, both molecularly and morphologically, as each species adapts to its environment \cite{Schneider:2016:EvolSpecializationTactilePerception, Valencia:2024:EvolutionSensoryReceptorsFamily}.
    The distributions of mechanoreceptors and their developmental mechanisms reported in anatomical and physiological studies can be regarded as outcomes of such adaptive processes.
    From this perspective, the design and comparison of sensory arrangements in this study can be interpreted as an attempt to experimentally reconstruct a subset of these possibilities and to quantitatively evaluate their functional characteristics.

    Traditionally, understanding why biological structures take particular forms has largely relied on anatomical comparisons and observational data.
    However, empirical methods for demonstrating the functional consequences of alternative configurations remain limited.
    It is inherently difficult to obtain sufficient data across a wide range of animal species, and, more critically, developmental and genetic constraints often prevent the realization of certain configurations within living organisms.
    To address this limitation, this study presents an experimental approach based on biomimetic robotics, in which configurations that are difficult or impossible to realize in biological systems can be physically reconstructed and their functional properties comparatively evaluated.

    By constructing physical systems and conducting repeated experiments, it becomes possible to obtain datasets of a scale that would be difficult to achieve using biological specimens, and to perform quantitative evaluations that take environmental variability into account.
    Previous studies have employed robotics to investigate biological mechanisms through physical reconstruction, such as work on the passive interlocking mechanism of the saltwater crocodile \cite{Ito:2023:CrocodylusPorosusMechanism}.
    In contrast, the present study not only reconstructs anatomically inspired configurations but also implements counterfactual configurations that are not observed in extant biological systems, and evaluates their functional characteristics.
    Furthermore, this study extends the scope from mechanical structures to sensory systems.
    Although it is not possible to fully replicate biological hardware, constructing systems with physical properties analogous to biological tissues enables complementary quantitative evaluation alongside traditional observational approaches.
    Using a biomimetic skin constructed at the tissue level, we confirmed that, for tactile tasks associated with human behavioral patterns, receptor arrangements similar to those in humans exhibited high performance.
    In addition, among the counterfactually constructed configurations, vibration-specialized arrangements were also identified, which are consistent with configurations observed in moles and waterfowl.
    From this perspective, this study presents an experimental framework that bridges biology and robotics, and the investigation of sensory organization can be regarded as one example of its application to understanding biological systems.

\section{Conclusion} \label{sec:self_organized_skin_conclusion}

    In this study, we constructed a biomimetic robotic skin in which multiple sensors corresponding to cutaneous receptors were embedded in diverse configurations, including arrangements that are difficult to realize in living systems, within compliant media mimicking skin tissue.
    We quantitatively evaluated these sensory arrangements through three passive tactile tasks: detection of weak contacts, a two-point discrimination-like assessment, and texture identification.
    As a result, configurations reflecting key characteristics of human receptor organization exhibited higher performance than the alternative arrangements examined.
    From a robotics perspective, these findings demonstrate that embedding sensors within compliant tissue-like media according to human-like sensory arrangements enables flexible design of multimodal tactile systems capable of discriminating pressure and texture, thereby providing new insights into principles of sensory system design.

    Furthermore, this study illustrates, using sensory arrangement as a case example, a biomimetic robotic approach that enables the physical implementation and comparison of biological body organizations and functional architectures that are difficult to investigate experimentally in living systems.
    The counterfactual sensory configurations examined here, although not observed biologically, can be regarded as part of a space of functionally realizable spatial organizations.
    Empirically comparing the functional consequences of sensory spatial organization, including such configurations, provides an experimental perspective for examining biological hypotheses concerning receptor arrangement.
    More broadly, this methodological framework may offer a complementary viewpoint for considering functional constraints and adaptive backgrounds related to body structure, and represents an example of how biomimetic robotics can be used to approach biological questions from an interdisciplinary perspective.

\funding
{
  This work was supported by JSPS KAKENHI Grant Numbers JP24KJ0697.
}

\roles
{
  A. Miki conceptualized the project, developed the methodology.
  S. Hasegawa assisted with the robot experiments.
  A. Miki conducted the investigation with support from R. Yoshimoto and Y. Sahara.
  A. Miki handled visualization and wrote the original draft.
  S. Hasegawa, K. Kawaharazuka and K. Okada did the reviewing the manuscript.
}

\section*{Ethical compliance}
{
  This study did not involve human participants or animals. Therefore, ethical approval and informed consent were not required.
}

\section*{Data Availability Statement}
{
  All data that support the findings of this study are included within the article (and any supplementary files).
}

\section*{Conflicts of Interest}
{
  The authors declare no conflicts of interest.
}

\suppdata
{
  \begin{figure}[htbp]
    \centering
    \includegraphics[width=1.0\linewidth]{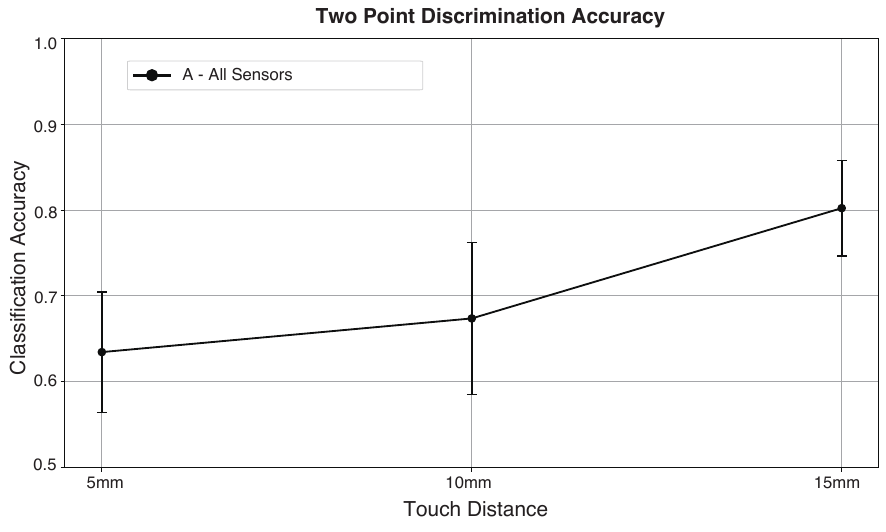}
    \caption{
      Classification accuracy for single- vs. multi-touch discrimination at varying inter-point distances (\SIrange{5}{15}{\milli\meter}) using Pattern A sensors.
      The curve is shown in black, solid line.
      Error bars indicate the standard deviation across trials.
      This corresponds to the Pattern A condition in the Main Figure (\figref{fig:hiro-two-point-discrimination-accuracy}).
    }
    \label{fig:hiro-two-point-discrimination-accuracy-with-error-bar-A-all}
  \end{figure}

  \begin{figure}[htbp]
    \centering
    \includegraphics[width=1.0\linewidth]{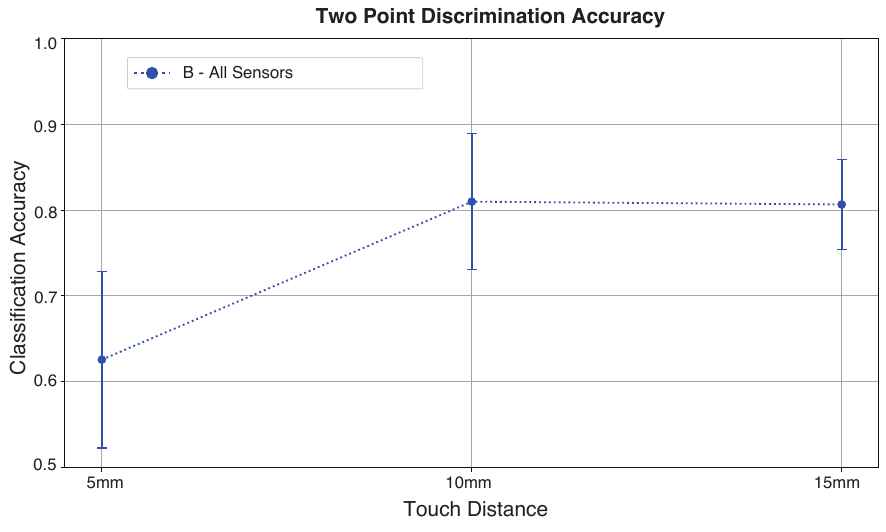}
    \caption{
      Classification accuracy for single- vs. multi-touch discrimination at varying inter-point distances using Pattern B (All Sensors).
      The curve is shown in blue, dotted line.
      Error bars indicate the standard deviation across trials.
      This corresponds to the Pattern B (All Sensors) condition in the Main Figure (\figref{fig:hiro-two-point-discrimination-accuracy}).
    }
    \label{fig:hiro-two-point-discrimination-accuracy-with-error-bar-B-all}
  \end{figure}

  \begin{figure}[htbp]
    \centering
    \includegraphics[width=1.0\linewidth]{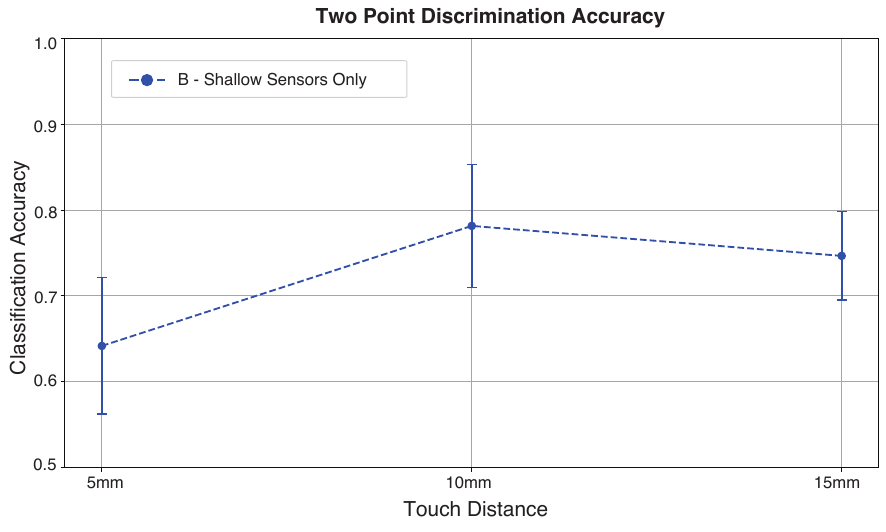}
    \caption{
      Classification accuracy for single- vs. multi-touch discrimination at varying inter-point distances using Pattern B (Shallow Sensors Only).
      The curve is shown in blue, dashed line.
      Error bars indicate the standard deviation across trials.
      This corresponds to the Pattern B (Shallow Sensors Only) condition in the Main Figure (\figref{fig:hiro-two-point-discrimination-accuracy}).
    }
    \label{fig:hiro-two-point-discrimination-accuracy-with-error-bar-B-shallow}
  \end{figure}

  \begin{figure}[htbp]
    \centering
    \includegraphics[width=1.0\linewidth]{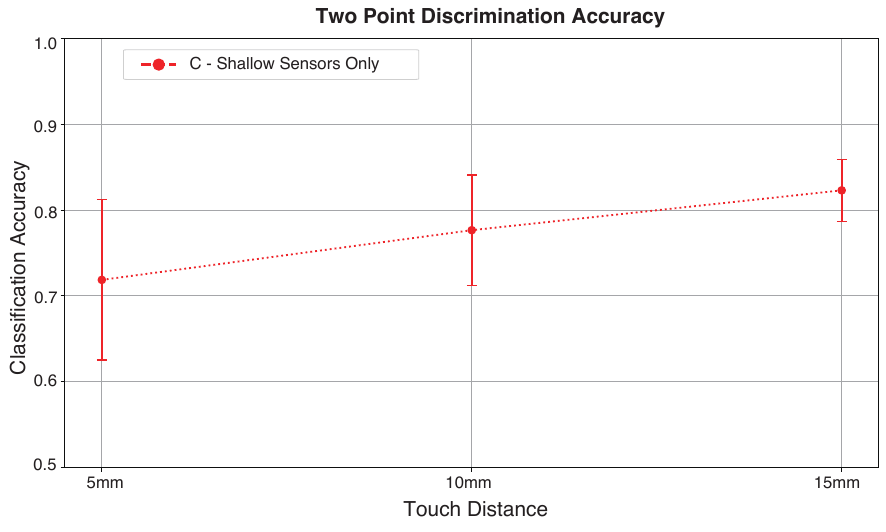}
    \caption{
      Classification accuracy for single- vs. multi-touch discrimination at varying inter-point distances using Pattern C (All Sensors).
      The curve is shown in red, dotted line.
      Error bars indicate the standard deviation across trials.
      This corresponds to the Pattern C (All Sensors) condition in the Main Figure (\figref{fig:hiro-two-point-discrimination-accuracy}).
    }
    \label{fig:hiro-two-point-discrimination-accuracy-with-error-bar-C-all}
  \end{figure}

  \begin{figure}[htbp]
    \centering
    \includegraphics[width=1.0\linewidth]{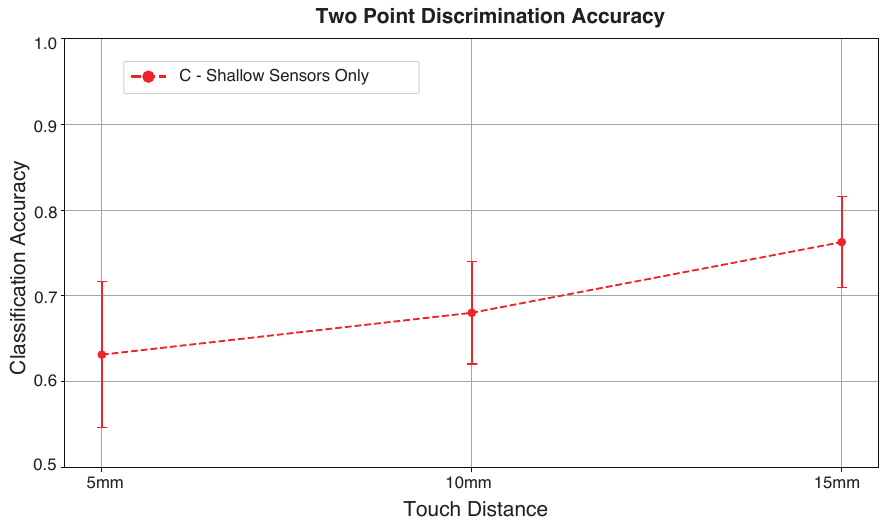}
    \caption{
      Classification accuracy for single- vs. multi-touch discrimination at varying inter-point distances using Pattern C (Shallow Sensors Only).
      The curve is shown in red, dashed line.
      Error bars indicate the standard deviation across trials.
      This corresponds to the Pattern C (Shallow Sensors Only) condition in the Main Figure (\figref{fig:hiro-two-point-discrimination-accuracy}).
    }
    \label{fig:hiro-two-point-discrimination-accuracy-with-error-bar-C-shallow}
  \end{figure}

  \begin{figure}[htbp]
    \centering
    \includegraphics[width=0.8\linewidth]{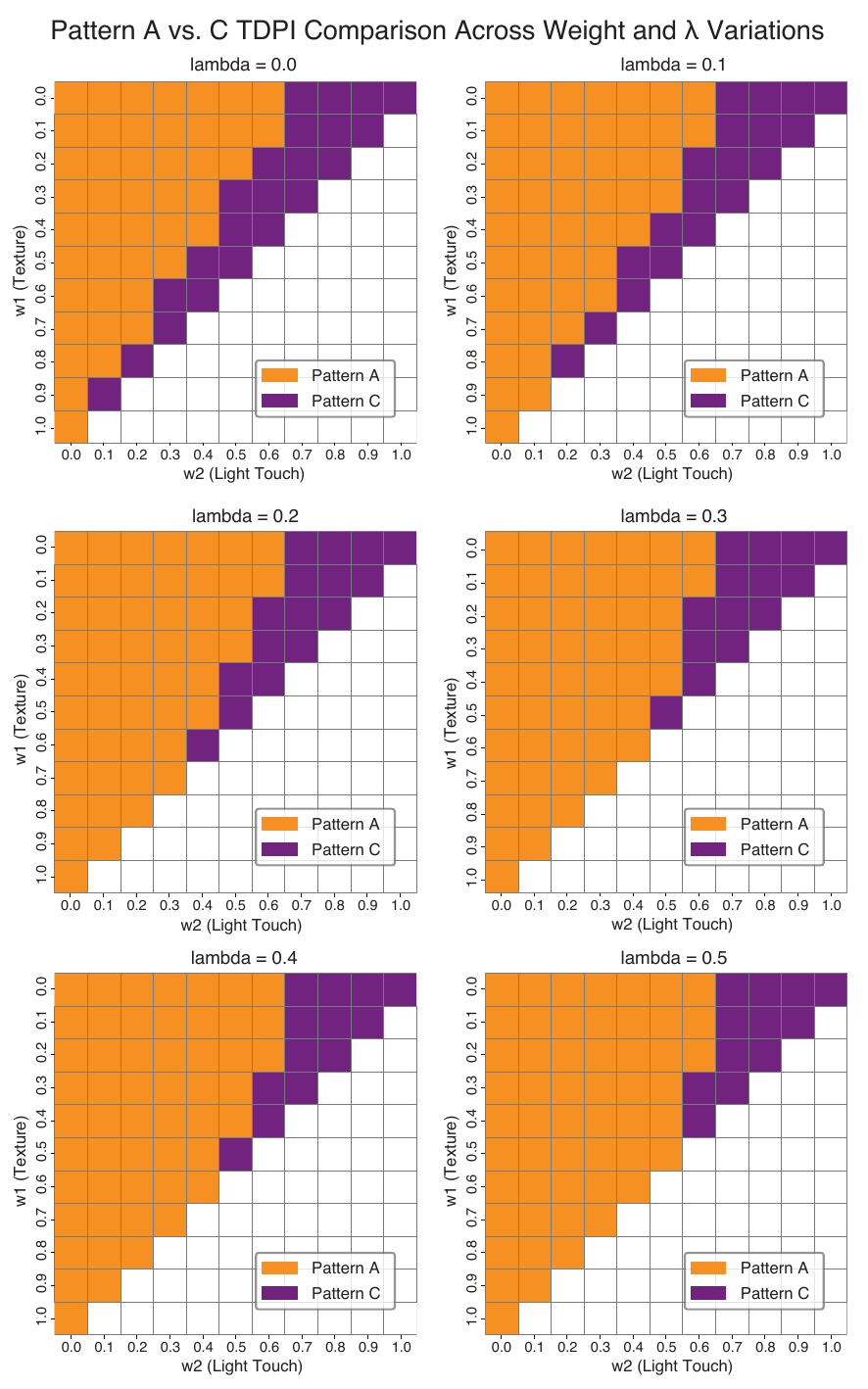}
    \caption{
      TDPI-based comparison between Pattern A and C under varying weight combinations and penalty coefficients $\lambda$.
      Each panel shows the preferred pattern (A or C) based on the TDPI for a fixed value of $\lambda$.
      The axes represent task weights: $w_1$ (vertical) for texture classification, $w_2$ (horizontal) for light touch detection, and $w_3 = 1 - w_1 - w_2$ for two-point discrimination.
      The results illustrate how the relative preference between Pattern A and C changes as the variability penalty coefficient $\lambda$ is varied.
    }
    \label{fig:ac-tdpi-comparison-by-lambda}
  \end{figure}
}

\FloatBarrier

\end{document}